\documentclass[runningheads,envcountsame]{llncs}
\usepackage{xcolor}

\usepackage{comment}

\usepackage{ecj,palatino,epsfig,latexsym,natbib}

\usepackage{amsmath, amsxtra, amsfonts, amssymb, amstext}
\usepackage{mathtools}
\usepackage{nicefrac}
\usepackage{xspace}
\usepackage[algo2e,ruled,vlined,linesnumbered]{algorithm2e}
\usepackage{algorithm}
\usepackage{algpseudocode}
\usepackage{multirow}
\usepackage{footnote}
\usepackage{hyperref}
\usepackage{thm-restate}
\usepackage{lineno}
\usepackage[shortlabels]{enumitem}
\usepackage{cases}
\usepackage{babel}
\usepackage{tikz}

\usepackage{graphicx}
\newcommand{\ooea}{$(1 + 1)$-EA\xspace}

\newcommand{\oclea}{$(1 , \lambda)$-EA\xspace}

\newcommand{\saolea}{\text{SA-}$(1, \lambda)$-EA\xspace}
\newcommand{\saoclea}{\text{SA-}$(1, \lambda)$-EA\xspace}

\newcommand{\srule}{$(1:s+1)$-rule\xspace}

\newcommand{\onemax}{\textsc{OneMax}\xspace}

\newcommand{\zeromax}{\textsc{ZeroMax}\xspace}
\newcommand{\OM}{\textsc{Om}\xspace}
\newcommand{\ZM}{Z}

\newcommand{\BinVal}{\textsc{BinVal}\xspace}
\newcommand{\Binval}{\textsc{BinVal}\xspace}
\newcommand{\dynBV}{\textsc{Dynamic BinVal}\xspace}
\newcommand{\DBV}{\textsc{DBv}\xspace}

\newcommand{\hottopic}{\textsc{HotTopic}\xspace}
\newcommand{\binary}{\textsc{Binary}\xspace}

\newcommand{\round}[1]{\ensuremath{\lfloor #1 \rceil}}
\newcommand{\linit}{\ensuremath{\lambda^{\mathrm{init}}}}

\newcommand{\xinit}{\ensuremath{x^{\mathrm{init}}}}
\newcommand{\pimp}{\ensuremath{p_{\mathrm{imp}}}}
\newcommand{\qimp}{\ensuremath{q_{\mathrm{imp}}}}

\newcommand{\EE}{\ensuremath{\mathbf{E}}}
\def\erw{{\ensuremath{\mathbf{E}}}}

\newcommand{\cF}{\ensuremath{\mathcal{F}}}

\DeclareMathOperator*{\argmax}{arg\,max}

\newcommand{\RR}{\ensuremath{\mathbb{R}}}
\newcommand{\NN}{\ensuremath{\mathbb{N}}}
\newcommand{\ZZ}{\ensuremath{\mathbb{Z}}}

\newcommand\numberthis{\addtocounter{equation}{1}\tag{\theequation}}

\spnewtheorem{fact}{Fact}{\bfseries}{\itshape}
\spnewtheorem{observation}[theorem]{Observation}{\bfseries}{\itshape}
\spnewtheorem{sclaim}{Claim}{\bfseries}{\itshape}
\newcommand{\proofitem}[1]{$\blacktriangleright$~\textbf{#1:}}

\newcommand\eps{\varepsilon}

\def\pr{{\mathbb P}}

\title{OneMax is not the Easiest Function for Fitness Improvements}

\ecjHeader{x}{x}{xxx-xxx}{201X}{OneMax is not the Easiest Function for Fitness Improvements}{M. Kaufmann, J. Lengler, M. Larcher and X. Zou}
\title{\bf OneMax is not the Easiest Function for Fitness Improvements}  

\author{\name{\bf Marc Kaufmann} \hfill \addr{marc.kaufmann@inf.ethz.ch}\\ 
        \addr{Department of Computer Science, ETH Zürich, 
        Zürich, Switzerland}
\AND
       \name{\bf Maxime Larcher} \hfill \addr{maxime.larcher@inf.ethz.ch}\\
        \addr{Department of Computer Science, ETH Zürich, 
        Zürich, Switzerland} 
\AND
       \name{\bf Johannes Lengler} \hfill \addr{johannes.lengler@inf.ethz.ch}\\
        \addr{Department of Computer Science, ETH Zürich, 
        Zürich, Switzerland}
\AND
        \name{\bf Xun Zou} \hfill \addr{xun.zou@inf.ethz.ch}\\
        \addr{Department of Computer Science, ETH Zürich, 
        Zürich, Switzerland}
}

\begin{document}

\maketitle

\begin{abstract}

We study the $(1:s+1)$ success rule for controlling the population size of the $(1,\lambda)$-EA. It was shown by Hevia Fajardo and Sudholt that this parameter control mechanism can run into problems for large $s$ if the fitness landscape is too easy. They conjectured that this problem is worst for the \onemax benchmark, since in some well-established sense \onemax is known to be the easiest fitness landscape. In this paper we disprove this conjecture. We show that there exist $s$ and $\eps$ such that the self-adjusting $(1,\lambda)$-EA with the \srule optimizes \onemax efficiently when started with $\eps n$ zero-bits, but does not find the optimum in polyno\-mial time on \dynBV. Hence, we show that there are landscapes where the problem of the \srule for controlling the population size of the \oclea is more severe than for \onemax. The key insight is that, while \onemax is the easiest function for decreasing the distance to the optimum, it is \emph{not} the easiest fitness landscape with respect to finding fitness-improving steps.

\end{abstract}


\section{Introduction}

The \onemax function assigns to a bit string $x$ the number of one-bits in $x$. Despite, or rather because of its simplicity, this function remains one of the most important unimodal benchmarks for theoretical analysis of randomized optimization heuristics, and specifically of Evolutionary Algorithms (EAs). A reason for the special role of this function is the result by% Doerr, Johannsen and Winzen
~\cite{doerr2012multiplicative} that it is the easiest function with a unique optimum for the \ooea in terms of expected optimization time. This result has later been extended to many other EAs by~\cite{sudholt2012new} and to stochastic dominance instead of expectations by~\cite{witt2013tight}. Easiest and hardest functions have become research topics of their own, e.g.\ see the work of~\cite{droste2006rigorous}, \cite{he2014easiest}, \cite{corus2017easiest} or~\cite{doerr2018optimal}.

Whether a benchmark is easy or hard is crucial for parameter control mecha\-nisms (PCMs); this was notably discussed by~\cite{Eiben99parameter} and~\cite{doerr2020theory}. 
Such mechanisms address the classical problem of setting the parameters of algorithms. They can be regarded as meta-heuristics which automatically tune the parameters of the underlying algorithm. The hope is that (i) optimization is more robust with respect to the meta-parameters of the PCM than to the parameters of the underlying algorithm, and (ii) PCMs can deal with situations where different optimization phases require different parameter settings for optimal performance~\citep{badkobeh2014unbiased,bottcher2010optimal,doerr2015black,doerr2020optimal,doerr2021runtime}.

To this end, PCMs often rely on an (often implicit) measure of how easy the optimization process currently is. One of the most famous examples is the \emph{$(1:s+1)$-rule} for step size adaptation in continuous optimization~\citep{kern2004learning,rechenberg1978evolutionsstrategien,devroye72compound,schumer1968adaptive}. It is based on the heuristic that improving steps are easier to find if the step size is small, but that larger step sizes are better at exploiting improvements, if improvements are found at all. Thus we have conflicting goals requiring small and large step sizes, respectively, and we need a compromise between those goals. The \srule resolves this conflict by defining a \emph{target success rate}\footnote{Traditionally the most popular value is $s=4$, leading to the famous \emph{one-fifth rule} originally proposed by~\cite{auger2009benchmarking}.} of $q_s = 1/(s+1)$, and increasing the step size if the success rate (the fraction of steps which find an improvement) is above $q_s$, and decreasing the step size otherwise. Thus it chooses larger step sizes in environments where improvements are easy to find, and chooses smaller step sizes in more difficult environments.

More recently, the \srule has been extended to parameters in discrete domains, in particular to the mutation rate by~\cite{doerr2018simple,doerr2021self} and offspring population size by~\cite{hevia2021arxiv,hevia2021self,hevia2021selfb}. \cite{doerr2018optimal} showed that for $s=4$, the Self-Adjusting $(1+(\lambda,\lambda))$-GA optimizes \onemax in expected linear time. For the self-adapting \oclea with the \srule, or \saolea for short, \cite{hevia2021arxiv,hevia2021self} showed an inter\-est\-ing collection of results on \onemax. They showed that optimization is highly efficient if the success ratio $s$ is less than one. In this case, the algorithm achieves optimal population sizes $\lambda$ throughout the course of optimization. The optimal $\lambda$ ranges from constant population sizes at early stages to almost linear (in the problem di\-men\-sion $n$) values of $\lambda$ for the last steps. On the other hand, the mechanism provably fails completely if $s\ge 18$.\footnote{Empirically they found the threshold between the two regimes to be around $s\approx 3.4$.} Then the algorithm does not even manage to obtain an $85\%$ approximation of the optimum in polynomial time. 

Hevia Fajardo and Sudholt also gave some important insights into the reasons for failure. The problem is that for large values of $s$, the algorithm implicitly targets a population size $\lambda^*$ with a rather small success rate.\footnote{See the previous work by the authors~\citep[Section~2.1]{kaufmann2022arxiv} for a detailed discussion of the target population size $\lambda^*$.} However, the \oclea is a non-elitist algorithm, i.e., the fitness of its population can decrease over time. This is particularly likely if $\lambda$ is small. So for large values of $s$, the PCM chooses population sizes that revolve around a rather small target value $\lambda^*$. It is still guaranteed that the algorithm makes progress in successful steps, which comprise a $\approx 1/(s+1)$ fraction of all steps. But due to the small population size, it loses performance in some of the remaining $\approx s/(s+1)$ fraction of steps, and this loss cannot be compensated by the gain of successful steps. 

A counter-intuitive aspect of this bad trade-off is that it only happens when success is too easy. If success is hard, then the target population size $\lambda^*$ is also large. In this case, the losses in unsuccessful steps are limited: most of the time, the offspring population contains a duplicate of the parent, in which case the loss is zero. Very recently, \cite{hevia2022hard} showed that for so-called \emph{everywhere hard} functions, where for all search points the probability of finding an improvement is bounded by $n^{-\varepsilon}$ for a constant $\varepsilon >0$, $\lambda$ quickly reaches a sufficiently large value such that a duplicate of the parent is guaranteed. So counter-intuitively, \emph{easy fitness landscapes lead to a high runtime}. For \onemax, this means that the problems do not occur close to the optimum, but only at a linear distance from the optimum. This result was implicitly shown by~\cite{hevia2021arxiv,hevia2021self} and explicitly by~\cite{kaufmann2022arxiv}: for every $s>0$ there is $\eps >0$ such that if the \saolea with success ratio $s$ starts within distance at most $\eps n$ from the optimum, then it is efficient with high probability, i.e., with probability $1-o(1)$. 

The results by~\cite{hevia2021arxiv,hevia2021self} were for \onemax, but~\cite{kaufmann2022arxiv,kaufmann2022self} could show that this result holds for all monotone, and even for all \emph{dynamic monotone functions}\footnote{A function $f:\{0,1\}^n \to \RR$ is monotone if flipping a zero-bit into a one-bit always improves the fitness. In the dynamic monotone setting, selection may be based on a different function in each generation, but it must always be a monotone function. The formal definition is not relevant for this paper, but can be found in precedent work by the authors~\citep{kaufmann2022arxiv}.}. Only the threshold for $s$ changes, but it is a universal threshold: there exist $s_1 > s_0 > 0$ such that for every $s<s_0$, the \saolea is efficient on \emph{every} (static or dynamic) monotone function, while for $s>s_1$ the \saolea fails for \emph{every} (static or dynamic) monotone function to find the optimum in polynomial time. Moreover, for all $s>0$ there is $\eps >0$ such that with high probability the \saolea with parameter $s$ finds the optimum of every (static or dynamic) monotone function efficiently if it starts at distance $\eps n$ from the optimum. Hence, all positive and negative results from~\cite{hevia2021arxiv} are not specific to \onemax, but generalize to every single function in the class of dynamic monotone functions. We also note that this class falls into more general frameworks of partially ordered functions, introduced by \cite{jansen2007brittleness} and further studied by \cite{colin2014monotonic}, that are easy to optimize under certain generic assumptions.

To summarize, small success rates (large values of $s$) are problematic, but only if the fitness landscape is too easy. Based on this insight, and on the aforementioned fact that \onemax is the easiest function for the \ooea, Hevia Fajardo and Sudholt conjectured that \onemax is the most problematic situation for the \saolea: ``given that for large values of $s$ the algorithm gets stuck on easy parts of the optimisation and that \onemax is the easiest function with a
unique optimum for the \ooea, we conjecture that any $s$ that is efficient on \onemax would also be a good choice for any other problem.'' In the terminology above, the conjecture says that the threshold $s_0$ below which the \saolea is efficient for all dynamic monotone functions, is the same as the threshold $s_0'$ below which the \saolea is efficient on \onemax. Note that the exact value of $s_0'$ is not known theoretically except for the bounds $1\le s_0' \le 18$, although Hevia Fajardo and Sudholt empirically showed that $s_0' \approx 3.4$. %~\cite{hevia2021arxiv}. 
If the conjecture was true, then experiments on \onemax could provide parameter control settings for the \saolea that work in much more general settings.

However, in this paper we disprove the conjecture. Moreover, our result makes it more transparent in which sense \onemax is the easiest benchmark for the \ooea or the \oclea, and in which sense it is not. As established by~\cite{doerr2012multiplicative} and \cite{witt2013tight}, it is the easiest benchmark in the sense that for no other function with a unique global optimum, the distance to the optimum decreases faster than for \onemax.
However, it is \emph{not} the easiest function in the sense that it is the easiest to make a fitness improvement, i.e., to find a successful step. Rephrased, other functions make it easier to find a fitness improvement than \onemax. For the problems of the \srule described above, the latter variant is the important one, since the \srule adjusts its population size based on the success probability of finding a fitness improvement.

\subsection{Our Result}

We are far from being able to determine the precise efficiency threshold $s_0'$ even in the simple setting of \onemax, and the upper and lower bound $1 \le s_0' \le 18$ are far apart from each other. Therefore, it is no option to just compute and compare the thresholds for different functions. Instead, we will identify a setting in which we can indirectly compare the efficiency thresholds for \onemax and for some other function, without being able to compute either of the thresholds explicitly. For this reason, we only study the following, rather specific setting that makes the proof feasible.

We show that there are $\eps >0$ and $s>0$ such that the \saolea with parameter~$s$ (and suitably chosen other parameters), started at any search point at distance exactly $\eps n$ from the optimum
\begin{itemize}
    \item finds the optimum of \onemax in $O(n)$ generations with high probability;
    \item does not find the optimum of \dynBV in polynomial time with high probability.
\end{itemize}
The definition of the \dynBV function can be found in Section~\ref{sec:benchmarks}. The key ingredient to the proof is showing that at distance $\eps n$ from the optimum, \dynBV makes it easier to find a fitness improvement than \onemax (Lemma~\ref{lem:pimp-relations}). Since \emph{easy} fitness landscapes translate into poor choices of the population size of the \saolea, and thus to \emph{large} runtimes, we are able to find a value of $s$ that separates the two functions: for this $s$ and for distance $\eps n$ from the optimum, the algorithm will have drift \emph{away} from the optimum for \dynBV (leading to exponential runtime), but drift \emph{towards} the optimum for \onemax. Since the fitness landscape for \onemax only gets harder closer to the optimum, we can then show that the drift remains positive all the way to the optimum for \onemax. A high-level sketch with more detail can be found in Section~\ref{sec:sketch}.

A limitation of our approach is that we start with a search point at distance~$\eps n$ from the optimum, instead of a uniformly random search point in $\{0,1\}^n$. This simplifies the calculations substantially, and it disproves the strong ``local'' inter\-pre\-ta\-tion of the conjecture of~\cite{hevia2021arxiv} that an $s$ that works for \onemax in some specific part of the search space also works in the same part for all other dynamic monotone functions. Our choice leaves open whether some weaker version of the conjecture of~\cite{hevia2021arxiv} might still be true. But since our argument refutes the intuitive foundation of the conjecture, we do not think that this limitation is severe.

Another limitation is that we use a dynamic monotone function instead of a static one. So we show that \onemax is not the easiest function in the class of dynamic monotone functions, but it could still be easiest in the smaller class of static monotone functions. Again, we have decided for this option for technical simplicity. We believe that our results for \dynBV could also be obtained with very similar arguments for a static \hottopic function as introduced by~\cite{lengler2018drift} and~\cite{lengler2019general}. However, \dynBV is simpler than \hottopic functions, and the dynamic setting allows us to avoid some technical difficulties. We thus restrict ourselves to experiments in Section~\ref{sec:simulations} for this hypothesis, and find that \onemax indeed has a harder fitness landscape (in terms of improvement probability) than other static monotone or even linear functions, and consistently (but counter-intuitive) the \saolea chooses a higher population size for \onemax. For some values of $s$, this leads to positive drift and efficient runtime on \onemax, while the same algorithm has negative drift and fails on other functions. 

Finally, apart from the success ratio $s$, the \saolea also comes with other parameters. For the \emph{mutation rate} we use the standard choice $1/n$; our analysis would also work for a mutation-rate $c/n$ for a constant $0<c<1$, with possibly different \(\eps\) and \(s\). The \emph{update strength} $F>1$ is the parameter that controls the speed at which the population size \(\lambda\) changes: in case of success we decrease \(\lambda\) by a factor of \(F\), and otherwise we increase it multiplicatively by \(F^{1/s}\). We refer to Section~\ref{sec:algorithm} for the details. A slight mismatch with the setup studied by~\cite{hevia2021arxiv} is that we choose $F = 1+o(1)$, while Hevia Fajardo and Sudholt focused on constant $F$. Again, this simplifies the analysis, but the restriction does not seem crucial for the conceptual understanding that we gain in this paper.

%\subsection{Related Work}

%The performance of the \oclea for static population sizes is well-understood~\cite{rowe2014choice,antipov2019efficiency}. %, in particular it is known that the optimum is only found efficiently if the population size is at least $C\log n$ for a suitable constant $C>0$. 

\section{Preliminaries and Definitions}\label{sec:prelim}

We use \(\log\) to denote the natural logarithm, i.e., \(\log e = 1\).
Our search space is always $\{0,1\}^n$. Throughout the paper we will assume that $s>0$  is independent of $n$ while $n\to \infty$, but \(F = 1 + o(1)\) will depend on \(n\). 
%, and we denote by $\supp\{x\} := \{i\in [n] \mid x_i = 1\}$ the \emph{support} of a bit string $x\in \{0,1\}^n$. 
We say that an event $\mathcal E = \mathcal{E}(n)$ holds \emph{with high probability} or \emph{whp} if $\Pr[\mathcal E] \to 1$ for $n\to\infty$. We will write $x=a \pm b$ as shortcut for $x \in [a-b,a+b]$. Throughout the paper we will measure drift \emph{towards} the optimum, so a positive drift always points towards the optimum, and a negative drift points away from the optimum. 
%We denote the negation of an event $\mathcal E$ by $\overline{\mathcal E}$, and by $\indicator{\mathcal E}$ the indicator of $\mathcal E$, i.e., $\indicator{\mathcal E} = 1$ if $\mathcal E$ holds and $\indicator{\mathcal E} = 0$ otherwise. 

\subsection{The Algorithm: \saolea}\label{sec:algorithm}

The \oclea is the algorithm that generates $\lambda$ offspring in each generation, and picks the fittest one as the unique parent for the next generation. All offspring are generated by standard bit mutation, where each of the $n$ bits of the parent is flipped independently with probability $1/n$. The performance of the \oclea for static population size $\lambda$ is well-understood, notably thanks to the work of~\cite{rowe2014choice} and~\cite{antipov2019efficiency}. 

We will consider the self-adjusting \oclea with $(1:s+1)$-success rule to control the population size $\lambda$, with success rate $s$ and update strength $F$, and we denote this algorithm by \saolea. It is given by the pseudocode in Algorithm~\ref{alg:saolea}. The key difference from the standard \oclea is that the population size \(\lambda\) is updated at each step: whenever a fitness improvement is found, the population is reduced to \(\lambda / F\) and otherwise the population is increased to \(\lambda F^{1/s}\). 
Note that the parameter $\lambda$ may take non-integral values during the execution of the algorithm; when this happens, the number of children generated is the integer \(\round{\lambda}\) closest to \(\lambda\). 

\begin{algorithm}
\caption{\saolea with success rate $s$, update strength $F$, mutation rate $1/n$, initial start point $\xinit\in\{0,1\}^n$ and initial population size $\linit = 1$ for maximizing a fitness function $f:\{0,1\}^n \to \RR$. \label{alg:saolea} }
\SetKwInput{Init}{Initialization}
\SetKwInput{Mut}{Mutation}
\SetKwInput{Sel}{Selection}
\SetKwInput{Opt}{Optimization}
\SetKwInput{Upd}{Update}
\label{alg:saocl}

\Init{Set $x^0= \xinit$ and $\lambda^0 := 1$\;}
\Opt{
	\For{$t = 0,1,\dots$}
		{
		\Mut{
			\For{$j \in \{1,\dots,\round{\lambda^{t}} \}$}
				{
				$y^{t,j}\leftarrow$ mutate$(x^t)$ by flipping each bit of $x^t$ independently with probability $1/n$\;
				}}
		\Sel{
			 Choose $y^t = \argmax\{f(y^{t,1}), \dots, f(y^{t,\round{\lambda}})\}$, breaking ties randomly\;}
		\Upd{\\
			 \algorithmicif\ {$f(y^t) > f(x^t)$}\quad \algorithmicthen\ $\lambda^{t+1} \leftarrow \max\{1, \lambda^t/F\}$; \quad
			 \algorithmicelse\ 
				$\lambda^{t+1} \leftarrow F^{1/s}\lambda^t$;\\
			 $x^{t+1}\leftarrow y^t$;
			}
		}
}
\end{algorithm}
One way to think about the \saolea is that for each search point $x$ it implicitly has a \emph{target population size} $\lambda^* = \lambda^*(x)$ such that, up to rounding, the probability to have success (the fittest of $\lambda^*$ offspring is strictly fitter than the parent) equals the \emph{target success rate} $1/(s+1)$. The \srule ensures that there is a drift towards $\lambda^*$: whenever $\round \lambda >\lambda^*$, then $\lambda$ decreases in expectation, and it increases for $\round \lambda <\lambda^*$, both on a logarithmic scale. We refer the reader to previous work by the authors~\citep[Section~2.1]{kaufmann2022arxiv} for a more detailed discussion.

For the results of this paper, we will specify $s>0$ as a suitable constant, the initial population size is $\linit = 1$, the initial search point has exactly $\eps n$ zero-bits for a given $\eps$, and the update strength is $F= 1+\eta$ for some $\eta \in \omega(\log n/n) \cap o(1/\log n)$. %We will often omit the index $t$ if it is clear from the context.

\subsection{The Benchmarks: \onemax and \dynBV}\label{sec:benchmarks}
The first benchmark, the \onemax function, counts the number of one-bits \begin{align*}
    \OM(x) = \onemax(x)=\sum_{i=1}^n x_i.
\end{align*}
of $x\in\{0,1\}^n$. We also define the \zeromax function $\ZM(x) := n-\OM(x)$ as the number of zero-bits in $x$. 
%It measures the distance of $x$ from the global optimum $(1..1)$. 
Throughout the paper, we will denote its value at time $t$ by $Z^t := \ZM(x^t)$, omitting the superscript when the time is clear from context. 
% we also frequently use the scaling $\eps = Z/n$.
In this paper, \(\eps\) always denote a small constant and we seek to understand the behaviour of the algorithm when \(Z^t\) is approximately \(\eps n\). 

Our other benchmark is a \emph{dynamic} function, notably studied by~\cite{lengler2018noisy}. That means that in each genera\-tion $t$, we choose a different function $f^t$ and use $f^t$ in the selection update step of Algorithm~\ref{alg:saocl}. We choose \dynBV or \DBV, and which is the binary value function \Binval, applied to a randomly selected permu\-tation of the positions of the input string. This function has been used by~\cite{lengler2020large} and~\cite{lengler2021runtime} to model dynamic environments, or by~\cite{lehre2022more} to model uncertain objectives.
In detail, \Binval is the function that interprets a bit string as an integer representation and returns its value, so $\BinVal(x) = \sum_{i=1}^n 2^{i-1} \cdot x_{i}$. For the dynamic version, for each generation $t$ we draw uniformly at random a permutation $\pi^t$ of the set $\{1,\ldots,n\}$. The \DBV function for generation $t$ is then defined as 
\[\DBV^t(x)=\sum_{i=1}^n 2^{i-1} \cdot x_{\pi^t(i)}.\]
Note that in the update step, a re-evaluation of the parent $x^t$ with respect to $\pi^t$ is needed when comparing the fitness of $x^t$ and $y^t$. 

\subsection{Tools}
We will use drift analysis to analyze two random quantities: the distance $Z^t = \ZM(x^t)$ of the current search point from the optimum, and the population size $\lambda^t$ (or rather, $\log \lambda^t$). We notably use the following drift theorems to transfer results on the drift into expected hitting times. For an introduction to drift theory, we refer the interested reader to~\cite{lengler2020drift}. \par
% We will use drift analysis and notably the results of~\cite{lengler2020drift} to analyze two random quantities: The distance $Z^t = \ZM(x^t)$ of the current search point from the optimum, and the population size $\lambda^t$ (or rather, $\log \lambda^t$). We also use the following drift theorems to transfer results on the drift into expected hitting times.  \par
%and concentration results.
\begin{theorem}[Tail Bound for Additive Drift --- \cite{kotzing2016concentration}, Theorem~2]
    \label{thm:AdditiveDrift}
    Let \((X^t)_{t \ge 0}\) be a sequence of random variables over $\mathbb{R}$, each with finite expectation and let $n>0$. With $T=\min\{t\ge 0: X_t \ge n \mid X_0 \ge 0\}$, we denote the random variable describing the earliest point at which the random process exceeds $n$, given a starting value of at least~$0$. Suppose there are $\eps, c >0$ such that, for all $t$,
    \begin{enumerate}[(i)]
        \item $\EE[X^{t+1}-X^t\mid X^0,...,X^t, T > t] \ge \eps$, and
        \item $|X^t-X^{t+1}|<c$.
    \end{enumerate}
    Then, for all $s \ge \frac{2n}{\eps}$,
    \begin{align*}
        \Pr(T \ge s) \le \exp{\left(-\frac{s \eps^2}{8c^2}\right)}.
    \end{align*}
\end{theorem}

\begin{theorem}[Negative Drift Theorem --- \cite{oliveto2015improved}, Theorem~2]
% \label{thm:NegativeDriftWithScaling}
    \label{thm:NegativeDrift}
    Let \((X^t)_{t \ge 0}\) be a sequence of random variables over $\RR$. Suppose there exists an interval $[a, b] \subseteq \RR$ and, possibly depending on $\ell := b-a$, a drift bound $\varepsilon := \varepsilon(\ell) > 0$ as well as a scaling factor $r := r(\ell)$ such that for all $t \ge 0$ the following three conditions hold:
    \begin{enumerate}[(i)]
        \item $\erw{[X^{t} - X^{t+1} \mid X^0, \dots, X^t; a < X_t < b}] \ge \varepsilon$; \label{thm:itm:NegativeDrift-1}
        \item $\Pr[{\lvert X^{t+1} - X^t\rvert \ge jr \mid X_0, \dots, X^t; a < X^t}] \le e^{-j}$ for all $j \in \NN$; \label{thm:itm:NegativeDrift-2}
        \item $1 \le r^2 \le \varepsilon \ell/(132\log(r/\varepsilon))$. \label{thm:itm:NegativeDrift-3}
    \end{enumerate}
    Let \(T = \inf \{t > 0 \mid X_t \ge b\} \) be the the earliest point in time when the process exceeds \(b\). If \(X_0 \le a\) then 
    \begin{align*}
        \Pr[{T \le e^{\varepsilon \ell/(132r^2)}}] = O(e^{-\varepsilon \ell/(132r^2)}).
    \end{align*}
\end{theorem}

To switch between differences and exponentials, we will frequently make use of the following estimates, taken from Lemma~1.4.2 -- Lemma~1.4.8 in~\cite{doerr2020probabilistic}. 
\begin{lemma}\label{lem:basic} $\ $
\begin{enumerate}[(i)]
    \item \label{lem:itm:basic-1}For all $y\ge 1$ and $0\le x \le y$,
    \begin{align*}
        (1-1/y)^{y} \le 1/e \le (1-1/y)^{y-1} \quad \text{ and } \quad   (1-x/y)^{y} \le e^{-x} \le (1-x/y)^{y-x}.
    \end{align*}
    % \item \label{lem:itm:basic-2}For all $0\le x \le 1$,
    % \begin{align*}
    %     1-e^{-x} \ge x/2.
    % \end{align*}
    \item \label{lem:itm:basic-3}For all $0 \le x\le 1$ and all $y\ge1$,
    \begin{align*}
        \tfrac{xy}{1+xy} \le 1- (1-x)^y \le xy.
    \end{align*}
%If moreover $xy \le 1$, then 
%\begin{align*}
%(1-x)^y \ge \tfrac{1}{1+xy} \ge 1-\tfrac{xy}{2}.
%\end{align*}
\end{enumerate}
\end{lemma}

%Finally, we will also use standard Chernoff bounds.
%\begin{theorem}[Chernoff Bound~{\cite[Section~1.10]{doerr2020probabilistic}}] \label{thm:Chernoff}
%Let $X_1, \ldots, X_n$ be in-dependent random variables taking values in $[0,1]$. Let $X = \sum_{i = 1}^n X_i$ and \par 
%\noindent let $1\ge \delta \ge 0$. Then 
%\begin{align}
%  \Pr[X \ge (1+\delta) E[X]] \le \exp\bigg(-\frac{\delta^2 E[X]}{3}\bigg).\label{eq:probCMUeasy}
%  \end{align}
%\end{theorem}

%\input{technical}
% \input{efficient_onemax}
\section{Main proof}

We start this section by defining some helpful notation. Afterwards, we give an informal sketch of the main ideas, before we give the full proof.
\begin{definition}
    \label{def:delta-p}
    Consider the \saolea optimizing a dynamic function \(f = f^t\), and let $Z^t = \ZM(x^t)$. For all times $t$ and all $i\in \ZZ$, we define 
    \begin{align*}
        p^{f,t}_i := \Pr[Z^{t}-Z^{t+1} = i \mid x^t, \lambda^t] \qquad \text{ and } \qquad
        \Delta^{f,t}_i := i \cdot p^{f,t}_i.
    \end{align*}
    We will often drop the superscripts \(f\) and \(t\) when the function and the time are clear from context. We also define \(p_{\ge i} := \sum_{j = i}^{\infty}{p_j}\) and \(\Delta_{\ge i} := \sum_{j=i}^\infty{\Delta_j}\); both \(p_{\le i}\) and \(\Delta_{\le i}\) are defined analogously. Finally, we write
    \begin{align*}
        \Delta^{f,t}  := \EE[Z^{t}-Z^{t+1} \mid x^t, \lambda^t] = \sum\nolimits_{i=-\infty}^\infty \Delta^{f,t}_i.
    \end{align*}
\end{definition}

Note that $i>0$ and $\Delta >0$ correspond to steps/drift \emph{towards} the optimum, while and $i<0$ and $\Delta <0$ correspond to steps/drift \emph{away from} the optimum.

\begin{definition}[Improvement Probability, Equilibrium Population Size]\label{def:lambdastar}
    Let \(x \in \{0,1\}^n\) and \(f\) be a strictly monotone function. Let \(y\) be obtained from \(x\) by flipping every bit independently with probability \(1/n\). We define
    \begin{align*}
        \pimp^f(x) := \Pr[ f(y) > f(x) ] \qquad \text{ and }\qquad \qimp^f(x,\lambda) := 1- (1-\pimp^f(x))^\lambda,
    \end{align*}
    as the probability that respectively a single offspring or any\footnote{Note that we define \(\qimp^f\) as above \emph{for all} \(\lambda\), even non-integer ones. Naturally the interpretation as the probability of finding some improvement only makes sense when \(\lambda\) is a positive integer.} of \(\lambda\) independent offspring improves the fitness of \(x\). We also define the \emph{equilibrium population size} as
    \begin{align}\label{eq:def:lambdastar}
        \lambda^{*, f}(x,s) := \log_{(1 - \pimp^f(x))}\big(\tfrac{s}{1+s}\big).
    \end{align}
    As the two functions we consider are symmetric (i.e.\ all bits play the same role) $\pimp^f$ only depends on $\ZM(x) = \eps n$ so in a slight abuse of notation we sometimes write $\pimp^f(\eps n)$ instead of $\pimp^f(x)$, and sometimes we also drop the parameters by writing just $\pimp^f$ when they are clear from the context. 
    
    Similarly we sometimes write $\qimp^f$ and $\lambda^{*,f}$.
\end{definition}

\begin{remark}\label{rem:equilibrium}
    For all \(x,s\) and \(f\), $\lambda^{*,f}(x,s)$ is chosen to satisfy 
    \begin{align*}
        \qimp^f(x,\lambda^{*,f}(x,s)) = \tfrac{1}{s+1},
    \end{align*}
    and is not necessarily an integer. Moreover, rounding $\lambda$ to the next integer can change the success probability by a constant factor. Thus we must take the effect of rounding into account. Fortunately, as we will show, the effect of changing the function $f$ from \onemax to \DBV is much larger than such rounding effects. 
\end{remark}

\subsection{Sketch of the Proof}\label{sec:sketch}

We have three quantities that depend on each other: the target population size $\lambda^*$, the target success rate $1/(s+1)$ and the distance $\eps := Z/n$ of the starting point from the optimum. Essentially, choosing any two of them determines the third one. In the proof we will choose $\lambda^*$ and $s$ to be large, and $\eps$ to be small. As $\eps$ is small, it is very unlikely to flip more than one zero-bit and the positive contribution to the drift is dominated by the term $\Delta_1$ (Lemma~\ref{lem:drift-bounds}~\ref{lem:itm:drift-bounds-2}). 
For \onemax we are also able to give a tight estimation of \(\Delta_{\le -1}\): for \(\lambda\) large enough we can guarantee that \(|\Delta_{\le -1}| \approx (1 - e^{-1})^\lambda\) (Lemma~\ref{lem:drift-bounds}~\ref{lem:itm:drift-bounds-3}).
% For \onemax, we also have the other direction: if $\lambda$ is large, then the negative contribution to the drift is dominated by $\Delta_{-1}$ (Lemma~\ref{lem:approximate_negative_drift}). 

The key to the proof is that under the above assumptions, the improvement probability $\pimp$ for \DBV is by a constant factor larger than for \onemax (Lemma~\ref{lem:pimp-relations}). Since it is unlikely to flip more than one zero-bit, the main way to improve the fitness for \onemax is by flipping a single zero-bit and no one-bits. Likewise, \DBV also improves the fitness in this situation. However, \DBV may also improve the fitness if it flips, for example, exactly one zero-bit and one one-bit. This improves the fitness if the zero-bit has higher weight, which happens with probability $1/2$. This already makes $\pimp^{\DBV}$ by a constant factor larger than $\pimp^{\OM}$. (There are actually even more ways to improve the fitness for \DBV.) As a consequence, for the same values of $s$ and $\eps$, the target population size $\lambda^*$ for \DBV is by a constant factor smaller than for \onemax (Lemma~\ref{lem:pimp-relations}~\ref{lem:itm:pimp-relations-3}).

This enables us to (mentally) fix some large $\lambda$, choose $\eps$ such that the drift for \onemax at $Z=\eps n$ is slightly positive (towards the optimum) and choose the value of $s$ that satisfies $\lambda^{*,\OM}(\eps n,s) = \lambda$. Here, `slightly positive' means that $\Delta_1 \approx 4|\Delta_{\le-1}|$. This may seem like a big difference, but in terms of $\lambda$ it is not. Changing $\lambda$ only affects $\Delta_1$ mildly. But adding just a single child (increasing $\lambda$ by one) reduces $\Delta_{\le-1}$ by a factor of $\approx 1 - 1/e$, which is the probability that the additional child is not a copy of the parent. So our choice of $\lambda^{*, \OM}, \eps$ and $s$ ensures positive drift for \onemax as long as $\lambda^t \ge \lambda^{*, \OM}-1$, but not for a much wider range. However, as we show, $\lambda^t$ stays concentrated in this small range due to our choice of $F=1+o(1)$. This already would yield progress for \onemax in a small range around $Z =\eps n$. To extend this to all values $Z \le \eps n$, we consider the potential function $G^t = Z^t - \tfrac{1}{2} \log_F(\lambda^t)$, and show that this potential function has drift towards zero (Corollary~\ref{cor:OM_positive_drift}) whenever $Z >0$. This part is similar to the arguments of~\cite{hevia2021arxiv} or~\cite{kaufmann2022arxiv}. For $\DBV$, we show that $\lambda^t$ stays in a range below $\lambda^{*,\DBV}(\eps n,s) + 1$, which is much smaller than $\lambda^{*,\OM}(\eps n,s)$, and that such small values of $\lambda^t$ give a negative drift on $Z^t$ (away from the optimum, Lemma~\ref{lem:construction_inefficient_DBV}). Hence, the algorithm is not able to cross the region around $Z = \eps n$ for \DBV.

\subsection{Full Proof}

In the remainder of this section we give the full proof, which follows the intuitive arguments presented above. In particular, we derive some relations between \(\lambda, s, \eps\) to find a suitable triple. Those relations and constructions only hold if the number of bits \(n\) is large enough. For instance, we wish to start at a distance \(\eps n\) from the optimum, meaning we need \(\eps n\) to be an integer. In all following statements, we implicitly assume that  $\eps n$ is a positive integer. In particular, this implies\footnote{In Lemma~\ref{lem:pimp-relations}~\ref{lem:itm:pimp-relations-2} and Lemma~\ref{lem:construction_inefficient_DBV} we consider a constant $\eps > 1/n$, introduce an \(\eps' = \eps'(\eps)\) and look at all states in the range \((\eps \pm \eps')n\). Again, we implicitly assume that \((\eps- \eps')n\) and \((\eps+ \eps')n\) are integers, since we use those in the calculations.} $\eps \ge 1/n$.

We start with the following lemma whose purpose is twofold. On the one hand it gives useful bounds and estimations of the probabilities of improvement; on the other hand it compares those probabilities of improvement for \onemax and \dynBV. In particular, the success probability for \onemax is substantially smaller than for \dynBV, meaning that \dynBV is easier than \onemax with respect to fitness improvements.

\begin{lemma}
    \label{lem:pimp-relations}
    Let $f$ be any dynamic monotone function and $1/n \le \eps \le 1$.
    \begin{enumerate}[(i)]
        \item Then $\pimp^f(\eps n) \le \eps$. More specifically for $\OM$ and $\DBV$, for all $n\ge 10$ we have%
        \begin{align*}%
            \pimp^{\OM}(\varepsilon n) = e^{-1}\varepsilon \pm 10\varepsilon^2 \qquad \text{and} \qquad \pimp^\DBV(\varepsilon n) = (1-e^{-1}) \varepsilon \pm 10\varepsilon^2.%
        \end{align*}\label{lem:itm:pimp-relations-general}%
    \end{enumerate}%
    In particular, there exist \(c > 0\) independent of \(n\) such that the following holds.
    \begin{enumerate}[(i)]
        \setcounter{enumi}{1}
        \item For every \(\delta \le 1 \le \lambda \le c \delta / \eps\), and every dynamic monotone function~$f$ we have \[\qimp^f(\eps n, \lambda) = (1 \pm \delta) \lambda \pimp^f (\eps n).\] \label{lem:itm:pimp-relations-1}
        \item There exists \(s_0 \ge 1\) independent of \(n\) such that the following holds. For every \(s \ge s_0\), every constant \(1/n<\varepsilon \le c\), and $f \in \{\OM, \DBV\}$ there exists a constant \(\eps'>0\) such that for all \(|x| \le \eps' n\) 
        \[
            \left| \lambda^{*, f}(\varepsilon n + x, s) - \lambda^{*, f}(\eps n, s) \right| \le 1/4.
        \] \label{lem:itm:pimp-relations-2}
        % OLD VERSION
        % \item For every \(s \ge 1\), every constant \(0<\varepsilon \le c\), and every dynamic monotone function $f$ there exists a constant \(\eps'>0\) such that  \[\lambda^{*, f}((\varepsilon - \eps')n, s) - \lambda^{*, f}((\varepsilon + \eps')n, s) \le 1/4.\] \label{lem:itm:pimp-relations-2}
        \item For every \(1/n \le \varepsilon \le c\) and \(s > 0\) we have \[0.5  \lambda^{*,\OM}(\varepsilon n, s) \le \lambda^{*, \DBV}(\varepsilon n, s) \le 0.6 \lambda^{*, \OM}(\varepsilon n , s).\] \label{lem:itm:pimp-relations-3}
    \end{enumerate}
\end{lemma}

\begin{proof}
    We start with \ref{lem:itm:pimp-relations-general}, and more precisely with the upper bound on \(\pimp^f(\eps n)\).
    For a general \emph{monotone} \(f\), the fitness can only improve if at least one zero-bit is flipped. Since the $i$-th zero-bit flips with probability $1/n$, by a union bound over the $\eps n$ zero-bits we have $\pimp^f(\eps n) \le \eps$.

    We now give more precise bounds for \(f \in \{\OM, \DBV\}\). As probabilities are necessarily lower and upper bounded by \(0\) and \(1\), the next statements are trivial when\footnote{Actually, it trivially holds for a wider range of values \(\eps \ge \tfrac{1}{\sqrt{10}} \approx \tfrac{1}{3.162...}\), but cutting at \(1/2\) is sufficient for the rest of our analysis.} \(1/2 \le \eps \le 1\), so we may assume \(\eps \le 1/2\).
    Let $A_1$ be the event of flipping exactly one zero-bit and an arbitrary number of one-bits,  \(A_{1,0}\) the sub-event of flipping exactly one zero-bit and no one-bits, and $A_{\ge 2}$ the event of flipping at least two zero-bits. Let $q_1 := \pr(A_1)$, $q_{1,0} := \pr(A_{1,0})$ and $q_{\ge 2} := \pr(A_{\ge 2})$ be their respective probabilities; we may lower and upper bound them as follows.
    We have $q_1 = \eps(1-1/n)^{\eps n -1}$. By Lemma~\ref{lem:basic}~\ref{lem:itm:basic-3} applied with \(x = 1/n\) and \(y = \eps n - 1\), we may bound $(1-1/n)^{\eps n -1} \ge 1- (\eps n-1)/n \ge 1-\eps$, which gives $\eps-\eps^2 \le q_1 \le \eps$. For $q_{1,0}$, we have \(q_{1,0} = \varepsilon (1 - 1/n)^{n-1}\). Using Lemma~\ref{lem:basic}~\ref{lem:itm:basic-1} with \(y = n\), we find \(e^{-1} \le (1 - 1/n)^{n-1} \le e^{-1} e^{1/n} \le e^{-1}(1 + 2/n)\le e^{-1}+\eps\) for all \(n \ge 1\), which gives $e^{-1}\eps \le q_{1,0} \le e^{-1}\eps + \eps^2$. Finally, for $q_{\ge 2}$, any fixed pair of two zero-bits has probability $1/n^2$ to be flipped, so by a union bound over all $\binom{\eps n}{2}$ such pairs we have $q_{\ge 2} \le \eps^2/2$. Summarizing, we have shown
    \begin{align*}
        q_{1} = \varepsilon \pm \eps^2 \qquad \text{and} \qquad q_{1,0} = e^{-1} \eps \pm \eps^2 \qquad \text{and} \qquad q_{\ge 2} \le \eps^2/2.
    \end{align*}
    For \onemax, the fitness always improves in the case $A_{1,0}$, may or may not improve in the case $A_{\ge 2}$, and does not improve in any other case. Hence,  $q_{1,0} \le \pimp^{\OM} \le q_{1,0} + q_{\ge 2}$, which implies the claim for \onemax.
    
    For \dynBV, we have to work a bit harder to compute the improve\-ment probability in the case of $A_1$. Consider this case, i.e., assume that exactly one zero-bit and an undetermined number of one-bits is flipped. Now we sort this zero-bit together with the $(1-\eps)n$ one-bits decreasingly by weight, thus obtaining a list of $(1-\eps)n +1$ bits. The zero-bit is equally likely to take any position $i\in [(1-\eps)n +1]$ in this list. Then the offspring is fitter than the parent if and only if none of the $i-1$ one-bits to the left of $i$ are flipped, which has probability $(1-1/n)^{i-1}$. Hence, conditional on $A_1$, the probability $q^\DBV_{|1}$ of improvement is
    \begin{align}\label{eq:lem:pimp01}\begin{split}
        q^\DBV_{|1} & = \frac{1}{(1-\eps)n+1}\sum_{i=1}^{(1-\eps)n+1} (1-1/n)^{i-1} %\\
        = \frac{1}{(1-\eps)n+1}\cdot
        \frac{1-(1-1/n)^{(1-\eps)n+1}}{1/n} \\
        & = \frac{1}{(1-\eps) + 1/n}\left(1-(1-1/n)^{(1-\eps)n+1}\right),
    \end{split}\end{align}
    where in the second step we used the formula for the geometric sum, $\sum_{i=1}^k x^{i-1} = (1-x^k)/(1-x)$. We separately estimate upper and lower bounds on this expression. For the first factor, we use $\tfrac{1}{(1-\eps)+1/n} \ge 1$ since $\eps \ge 1/n$, and $\tfrac{1}{(1-\eps)+1/n} \le \tfrac{1}{1-\eps} \le 1+2\eps$, which holds for $\eps \le 1/2$. For the second factor, we use Lemma~\ref{lem:basic}~\ref{lem:itm:basic-1} with \(y=n\) to obtain
    \[(1-1/n)^{(1-\eps)n+1} \le e^{-(1-\eps) -1/n} \le e^{-1}e^{\eps}\le e^{-1}(1+2\eps)\le e^{-1}+\eps \]
    and 
    \[(1-1/n)^{(1-\eps)n+1} \ge (1 - 1/n)^{n} \ge e^{-1}(1-1/n) \ge e^{-1} - \eps,\]
    where the first and last steps hold since $\eps\ge 1/n$. Plugging these bounds into~\eqref{eq:lem:pimp01} (mind the ``$1-$'' in the bracket of~\eqref{eq:lem:pimp01}) yields
    \begin{align*}
        1-e^{-1}-\eps \quad \; \le \; \quad q^\DBV_{|1} \; \le \; (1+2\eps)(1-e^{-1}+\eps) \le 1-e^{-1} +5\eps.
    \end{align*}
    Now we may use $q_1 \cdot q^\DBV_{|1} \le \pimp^{\DBV} \le q_1 \cdot q^\DBV_{|1} + q_{\ge 2}$, where
    \[
        q_1 \cdot q^\DBV_{|1} = (\eps \pm \eps^2)(1-e^{-1}\pm 5 \eps) = (1-e^{-1})\eps \pm [5 + (1 -e^{-1}) + 5\eps] \eps^2 = (1-e^{-1})\eps \pm 9 \eps^2,
    \]
    since we assume \(\eps \le 1/2\).
    Together with $q_{\ge 2} \le \eps^2/2$, this proves the claim for $\pimp^{\DBV}$.
    
    We now focus on \ref{lem:itm:pimp-relations-1}. Given an arbitrary monotone function \(f\), recall that \(\qimp^f = 1 - (1 - \pimp^f)^\lambda\), by Lemma~\ref{lem:basic}~\ref{lem:itm:basic-3} applied with \(x = \pimp^f\) and \(y = \lambda\) this can be bounded by
    \begin{align*}
        \frac{\lambda \pimp^f}{1 + \lambda \pimp^f} \; \le \; \qimp^f \; \le \; \lambda \pimp^f. 
    \end{align*}
    Since \(\pimp^f\le \eps\), if we choose \(\eps \le c \delta/\lambda\) for some absolute constant \(c\) small enough we indeed get \(\qimp^f = (1 \pm \delta) \lambda \pimp^f\) for every dynamic monotone function \(f\).
    
    The third result \ref{lem:itm:pimp-relations-2} is a consequence of the definition
    \begin{align*}
        \lambda^{*,f}(\eps n, s)
            &= \log_{(1 - \pimp^f)}(s/(s+1)) \; = \; \frac{\log(s/(s+1))}{\log(1 - \pimp^f(\eps n))}.
    \end{align*}    
    We know that \(\pimp^f (\eps n) \le \eps\), and since \(\eps \le c\) with \(c\) chosen smaller than \(1\), we may assume \(\pimp^f \le 1/2\). In particular, using \(-x - x^2 \le \log(1 - x) \le -x\) whenever \(x \le 1/2\), we find that \(\log(1 - \pimp^f) = -\pimp^f \pm (\pimp^f)^2 = -(1 \pm \eps) \pimp^f\). This gives
    %where \(c\) can be chosen arbitrarily small.  \(\log(1 - \pimp^f) = 1 \pm C \eps\) for some (potentially large, but absolute) constant \(C\). In particular, this implies
    \[
        \lambda^{*,f}(\eps n, s) \; = \; \frac{\log(s/(s+1))}{-(1 \pm \eps)\pimp^f(\eps n)}.
    \]
    When \(f \in \{\OM, \DBV\}\), we have established above that \(\pimp^{f}(\eps n) = C^f \eps \pm 10 \eps^2\) with \(C^\OM = e^{-1}\) and \(C^\DBV = 1 - e^{-1}\). Let us choose \(\eps' = \eps^2\): by changing the number of \(0\)-bits by \(x\) with \(|x| \le \eps' n\), we may change the value of \(\pimp^f\) by at most \(C\eps^2 \) where \(C = C^f + 20\) is an absolute constant. In particular, we find 
    \begin{align*}
        \lambda^{*,f}(\eps n + x, s) \; = \; \frac{\log(s/(s+1))}{-(1 \pm (\eps + \eps'))(\pimp^f(\eps n) \pm C\eps^2)} \; = \; \frac{\log(s/(s+1))}{-(1 \pm C' \eps)\pimp^f(\eps n)}
    \end{align*}
    for some (possibly large, but absolute) constant \(C'\). Here we used once again the fact that \(\pimp^f(\eps n)\) is of order \(\eps\) for both functions \(f \in \{\OM, \DBV\}\) that we consider. The difference \( | \lambda^{*,f}(\eps n + x, s) - \lambda^{*,f}(\eps n, s) |\) that we seek to bound may thus be at most 
    \begin{align*}
        \frac{\log(1 + 1/s)}{\pimp^f(\varepsilon n)} \cdot \left( \frac{1}{1 - C'\eps} - 1 \right) \; \le \; \frac{1}{s \cdot \pimp^f(\eps n)} \cdot \frac{C' \eps}{1 - C'\eps},
    \end{align*}
    where in the last step we used \(\log (1 + 1/s) \le 1/s\).
    For both functions we consider, we have proved \(\pimp^f(\eps n) \ge e^{-1} \eps - 10\eps^2\); this and the fact that \(\eps \le c\) with \(c\) as small as desired, we see that the difference above is at most \(4eC'/s\). Setting \(s_0 = 16 e C'\) then guarantees that we have \(|\lambda^{*,f}(\eps n + x, s) - \lambda^{*,f}(\eps n, s) | \le 1/4\).

    % Since \(s \ge 1\), changing \(\varepsilon\) by \(\varepsilon' = \varepsilon^2\) may change \(\lambda^*\) by \(C'\eps \cdot \log 2\), where \(C'\) is again a large but absolute constant. Up to possibly choosing \(c\) smaller, this quantity is bounded by \(1/4\).
    
    Lastly, we prove \ref{lem:itm:pimp-relations-3}. Again from the definition of \(\lambda^*\) we have 
    \begin{align*}
        \frac{\lambda^{*, \DBV}(\eps n,  s)}{\lambda^{*, \OM}( \eps n,  s)} \;
            &= \; \frac{\log_{(1 - \pimp^{\DBV}( \eps n))}\big(\tfrac{s}{1+s}\big)}{\log_{(1 - \pimp^{\OM}( \eps n ))}\big(\tfrac{s}{1+s}\big)} \; = \; \frac{\log(1-\pimp^{\OM}( \eps n))}{\log(1-\pimp^{\DBV}( \eps n))} \\
            &= \; \frac{\log( 1 - e^{-1} \eps \pm 2 \eps^2 )}{\log( 1 - (1-e^{-1}) \eps \pm 10 \eps^2 )}.
            % &= \frac{\pimp^\OM(\varepsilon n) + O(\varepsilon^2)}{\pimp^\DBV(\varepsilon n) + O(\varepsilon^2)}= \frac{2}{3} + O(\varepsilon).
    \end{align*}
    Since $\log(1-x) \approx -x$ for $x\to 0$, the above ratio tends to $e^{-1}/(1-e^{-1}) = 1/(e-1) = 0.58...$ as $\eps$ tends to zero. In particular, since \(\varepsilon < c\) is assumed small enough, the ratio is at least \(0.5\) and at most \(0.6\). \qed
%    
%     \begin{align*}
%         \frac{\lambda^{*, \DBV}(\eps n,  s)}{\lambda^{*, \OM}( \eps n,  s)} 
%             &= \frac{\log_{(1 - \pimp^{\DBV}( \eps n))}\big(\tfrac{s}{1+s}\big)}{\log_{(1 - \pimp^{\OM}( \eps n ))}\big(\tfrac{s}{1+s}\big)} = \frac{\log(1-\pimp^{\OM}( \eps n))}{\log(1-\pimp^{\DBV}( \eps n))}
%             % &= \frac{\pimp^\OM(\varepsilon n) + O(\varepsilon^2)}{\pimp^\DBV(\varepsilon n) + O(\varepsilon^2)}= \frac{2}{3} + O(\varepsilon).
%     \end{align*}
%     Substituting the bounds for $\pimp^\DBV$ yields the following bounds for the ratio
%     \begin{align*}
%         \frac{\log( 1 - e^{-1} \eps - 5 \eps^2 )}{\log( 1 - \frac{2}{5}\eps ) } \ge \frac{\log(1-\pimp^{\OM}( \eps n))}{\log(1-\pimp^{\DBV}( \eps n))} \ge \frac{\log( 1 - e^{-1} \eps + 5 \eps^2 )}{\log( 1 - \eps ) } 
%     \end{align*}
%    
%   %
%     %$\pimp^\DBV$: For $\pimp^\DBV= \frac{e-1}{e}\eps \pm 2\eps^2\subset \frac{e-1}{e}\eps \pm 5\eps^2$, the ratio tends to $\frac{1}{e-1}\approx 0.58$. In particular, since \(\varepsilon < c\) is assumed small enough,  that ratio is at least \(0.5\) and at most \(0.6\)} \\
%     This ratio tends to \(2/3\) as \(\eps\) tends to \(0\). In particular, since \(\varepsilon < c\) is assumed small enough,  that ratio is at least \(0.6\) and at most \(0.7\).
\end{proof}

We follow with a lemma that gives estimates of the drift of \(Z\). In particular, the first statement is one way of stating that \onemax is the easiest function with respect to minimizing the distance from the optimum. We will only apply it with \(f = \DBV\), but we believe that the result is interesting enough to be mentioned. 

\begin{lemma}
    \label{lem:drift-bounds}
    There exists a constant \(c>0\) such that the following holds for every dynamic monotone function \(f\). For every \(\delta > 0\), 
    % there exists \(\lambda_0\) such that 
    the following holds. 
    \begin{enumerate}[(i)]
        \item For all $\lambda \ge 1$, all $x\in\{0,1\}^n$ and all $i \in \NN$ we have \[\Delta^{f}_{\ge i}(x,\lambda) \le \Delta^{\OM}_{\ge i}(x,\lambda) \qquad \text{and} \qquad |\Delta^f_{\le -i}(x,\lambda)| \ge |\Delta^{\OM}_{\le - i}(x,\lambda)|.\] \label{lem:itm:drift-bounds-1}
        \item For every integer \(\lambda \ge \lambda_0\) and all \(1/n \le \varepsilon \le \min\{ 1, \, c \delta/\lambda \}\) we have \[\Delta_{\ge 1}^f(\eps n,\lambda) \le (1 + \delta) \Delta_1^{\OM}(\eps n,\lambda).\] \label{lem:itm:drift-bounds-2}
        \item There exists \(\lambda_0 \ge 1\) independent of \(n\) such that for every integer \(\lambda \ge \lambda_0\) and all\footnote{When applying the lemma, we consider fixed \(\delta, \lambda\) and an arbitrarily large \(n\). In particular, the interval \([1/n, c\delta / \lambda]\) is non-empty.} \(1/n \le \varepsilon \le \min\{ 1, \, c \delta/\lambda\}\) we have \[|\Delta_{\le -1}^{\OM}(\eps n,\lambda)| = (1 \pm \delta)(1 - e^{-1})^{\lambda}.\] \label{lem:itm:drift-bounds-3}
    \end{enumerate}
    Note that in the second point the right hand side is the drift with respect to \onemax, not to $f$.
\end{lemma}

\begin{proof}
    We prove items in order and start with \ref{lem:itm:drift-bounds-1}. For any $j\in\ZZ$, the event $Z^{t+1} \ge Z^t - j$ can only happen for \onemax if $\ZM(y) \ge \ZM(x) - j$ holds for all offspring $y$ of $x$. This is true, because, when optimising \onemax, we pick the offspring which minimizes $\ZM(y)$. Conversely, whenever all offspring satisfy \(\ZM(y) \ge \ZM(x) - j\), the event \(Z^{t+1} \ge Z^t - j\) must hold, regardless of the function \(f\) considered. 
    %In this case, regardless of the selection of \(f\), the selected offspring also satisfies $\ZM(y) \ge \ZM(x) - j$. 
    This implies that $p_{\le j}^\OM \le p_{\le j}^{f}$ and $p_{\ge j}^\OM \ge p_{\ge j}^{f}$ for all $j \in \ZZ$, and any dynamic monotone function \(f\). In particular, for all \(i \ge 0\)
    \begin{align*}
        0 \; \ge \; \Delta_{\le -i}^{\OM} \; 
            &= \;  \sum\nolimits_{j=-\infty}^{-i}{-j p_{j}^{\OM}} \;
            = \; -i \sum\nolimits_{j=-\infty}^{-i}{ p_{j}^{\OM} } - \sum\nolimits_{j=-\infty}^{-i}{(j-i) p_{j}^{\OM}} \\
            &= \; -i \cdot p_{\le -i}^{\OM} - \sum\nolimits_{j=-\infty}^{-i-1} p_{\le j}^{\OM} \; 
            \ge \; -i \cdot p_{\le -i}^{f} - \sum\nolimits_{j=-\infty}^{-i-1} p_{\le j}^{f} \;
            = \; \Delta_{\le -i}^{f},
    \end{align*}
    and similarly
    \begin{align*}
        \Delta_{\ge i}^{\OM} &= i \cdot p_{\ge i}^{\OM} + \sum\nolimits_{j=i+1}^{\infty} p_{\ge j}^{\OM} \ge i \cdot p_{\ge i}^{f} + \sum\nolimits_{j=i+1}^{\infty} p_{\ge j}^{f}  = \Delta_{\ge i}^{f}.
    \end{align*}

    We now turn to the proof of \ref{lem:itm:drift-bounds-2}, starting with $f=\OM$. Consider an arbitrary offspring. For any \(i\) zero-bits, there is probability \(n^{-i}\) that they all flip to one-bits. By union bound over \(\binom{\varepsilon n}{i} \le (\varepsilon n)^i / i!\) such choices, the probability that at least \(i\) zero-bits are flipped is at most \(\varepsilon^i/i!\). 
    Now union bounding over all children we find 
    \begin{align*}
        p_{\ge i}^{\OM} = \Pr[Z^t-Z^{t+1} \ge i] \le \lambda \varepsilon^i / i!.
    \end{align*}
    In turn this implies 
    \begin{align*}
        \Delta^{\OM}_{\ge 2} \; = \; \sum_{i=2}^\infty{ i p_i^\OM} \; \le \; \sum_{i = 2}^\infty{ i p^{\OM}_{\ge i} } \; \le \; \sum_{ i=2 }^\infty \frac{\lambda \eps^{i}}{(i-1)!} \; = \; \lambda \eps (e^\eps - 1) \; \le \; 2 \lambda \varepsilon^2, \numberthis \label{eq:lem:itm:drift-bounds-2-1}
    \end{align*}
    where the last step follow from \(\eps \le 1\).
    As a single child flipping exactly a zero-bit and no one-bit implies that \(Z^t\) decreases, we must also have
    \begin{align*}
        \Delta^{\OM}_1 = p^{\OM}_{1} \ge \varepsilon (1 - 1/n)^{n-1} \ge e^{-1} \varepsilon, \numberthis \label{eq:lem:itm:drift-bounds-2-2}
    \end{align*}
    by Lemma~\ref{lem:basic}~\ref{lem:itm:basic-1} applied with \(y = n\). Combining both equations~\eqref{eq:lem:itm:drift-bounds-2-1} and \eqref{eq:lem:itm:drift-bounds-2-2} gives \(\Delta^{\OM}_{\ge 2} \le 2e \lambda \eps \Delta^{\OM}_1 \le 2 e c \delta \Delta^{\OM}_1\), since \(\eps \le \delta c / \lambda\). Choosing \(c \le 1/(2e)\) ensures that \(\Delta^{\OM}_{\ge 2} \le \delta \Delta^{\OM}_1\), which implies \(\Delta^{\OM}_{\ge 1} \le (1 + \delta) \Delta^{\OM}_1\).
    % We have shown $\Delta_{\ge 2}^{\OM} \le \delta \Delta_1^{\OM}$, 
    The corresponding bound for $\Delta_{\ge 1}^{f}$ follows from \ref{lem:itm:drift-bounds-1}.

   We finish with the proof of \ref{lem:itm:drift-bounds-3} and start with the lower bound. As we are considering the function \(\OM\), \(Z^t\) must increase if every child flips at least a one-bit and no zero-bit. %, then \(Z^t\) must increase. 
   Hence we have 
    \begin{align*}
        |\Delta^{\OM}_{\le -1}| 
            &\ge \left( (1 - (1 - 1/n)^{(1-\varepsilon) n}) \cdot (1 - 1/n)^{\varepsilon n} \right)^{\lambda}\\
            &= \left((1-1/n)^{\eps n} - (1 - 1/n)^n \right)^\lambda. % (1 - e^{-1} + O(\varepsilon))^\lambda.
    \end{align*}
    By Lemma~\ref{lem:basic}~\ref{lem:itm:basic-1} applied with \(y = n\), the first term in parentheses may be bounded from below by \((1-1/n)^{\eps n}  = \left((1 - 1/n) \cdot (1 - 1/n)^{n-1} \right)^\eps \ge (1-1/n)^\eps e^{-\eps}\); since \(\eps \ge 1/n\), we have \((1 - 1/n)^\eps \ge (1 - \eps)^\eps \ge 1-\eps\) and we also have \(e^{-\eps} \ge 1-\eps\) by Lemma~\ref{lem:basic}~\ref{lem:itm:basic-1} with \(y = 1/\eps\). The second term of the display above may simply be handled by Lemma~\ref{lem:basic}~\ref{lem:itm:basic-1}: \((1-1/n)^n \le e^{-1}\). Combining everything we find 
    \begin{align*}
        |\Delta^{\OM}_{\le -1}| 
            &\ge \left((1-\eps)^2 - e^{-1}\right)^\lambda \ge (1 - e^{-1})^\lambda \cdot \left(1 - \tfrac{2}{1-e^{-1}} \cdot \eps\right)^\lambda \\
            &\ge (1 - e^{-1})^\lambda \cdot \left(1 - \tfrac{2\lambda}{1-e^{-1}}\cdot \eps\right),
    \end{align*}
    where the last step follows from an application of Lemma~\ref{lem:basic}~\ref{lem:itm:basic-3} with \(x = \tfrac{2}{1-e^{-1}}\eps\) and \(y = \lambda\). 
    Since \(c\) is assumed to be sufficiently small and \(\varepsilon < c \delta / \lambda\), we have \(|\Delta_{\le -1}| \ge (1 - \delta)(1 - e^{-1})^\lambda\).

    The lower bound is proved and we now focus on the upper bound. Since we may express \(|\Delta^{\OM}_{\le -1}| = \sum_{i \ge 1}{p^{\OM}_{\le -i}}\), it suffices to bound each term appearing in the sum. As we are considering the fitness function \onemax, the number of zero-bits \(Z^t\) increases by at least \(i\) only if every child flips at least \(i\) one-bits. The probability that one child leaves all \(1\)-bits unchanged is \((1 - 1/n)^{(1-\eps)n} \ge e^{-1}\) by Lemma~\ref{lem:basic}~\ref{lem:itm:basic-1} applied with \(y = n\) since \(\eps \ge 1/n\). In particular, we have 
    \begin{align*}
        p^{\OM}_{\le -1} 
        %\le \left( 1 - (1 - 1/n)^{(1-\varepsilon)n} \right)^\lambda 
        \le (1 - e^{-1})^\lambda.
    \end{align*}
    For \(i > 1\) and an arbitrary offspring, the probability of flipping \(i\) or more \(1\)-bits is at least \(\binom{(1-\eps) n}{i} (1/n)^i \le 1/i!\), so that 
    \begin{align*}
        p^{\OM}_{\le -i} 
        %\le \left(\binom{(1-\varepsilon)n}{i} n^{-i}\right)^\lambda 
        \; \le \; (i!)^{-\lambda} \; \le \; 2^{(1-i) \lambda}.
    \end{align*}
    This immediately implies 
    \begin{align*}
        |\Delta^{\OM}_{\le -1}| \;
            = \; p^{\OM}_{\le -1} + \sum_{i=2}^{\infty}{p^{\OM}_{\le -i}} \;
            \le \; (1 - e^{-1})^\lambda + 2^{1-\lambda}. %\numberthis \label{}
    \end{align*}
    The second term decays faster than the first since \((1-e^{-1}) \ge 2^{-1}\). 
    Therefore, for a sufficiently large \(\lambda_0\) --- of order at least \(\sim \log(1/\delta)\), which is a constant w.r.t.\ \(n\) ---  the second term of the right-hand side is at most \(\delta \cdot (1 - e^{-1})^\lambda\), and this finishes the proof of \ref{lem:itm:drift-bounds-3}.
    \qed
\end{proof}

\begin{remark}
    Observe that in Lemma~\ref{lem:pimp-relations} (resp.~\ref{lem:drift-bounds}), \(c\) only serves as an upper bound for \(\eps\), or as a scaling factor for such a bound. In particular, if Lemma~\ref{lem:pimp-relations} or Lemma~\ref{lem:drift-bounds} can be applied for some \(c\), then it can also be applied for all \(c' < c\). Consequently, we may always find a \(c\) sufficiently small so that both Lemmas~\ref{lem:pimp-relations} and~\ref{lem:drift-bounds} can be applied simultaneously; we use this observation several times in the rest of the proof below.
\end{remark}

We now come to the core of our proof, which is finding a suitable triple \(\lambda^*, \varepsilon, s\).

\begin{lemma}
    \label{lem:construction_equilibrium_OM}
    For every \(\delta \in (0, 1]\) there exists a \(\lambda_0 \ge 1\) constant with respect to \(n\) such that the following holds. For every integer \(\lambda \ge \lambda_0\), there exist \(\tilde{\varepsilon}, \tilde{s}\) depending 
    on \(\lambda\),
    but constant with respect to \(n\)
    such that 
    \begin{enumerate}[(i)]
        \item \(\lambda = \lambda^{*, \OM}(\tilde{\eps} n, \tilde{s})\); 
        \item and \(\Delta^{\OM}_{\ge 1}(\tilde \eps n, \lambda) = (4 \pm \delta) |\Delta^{\OM}_{\le -1}(\tilde \eps n, \lambda)|.\) % = (1 \pm \delta)/(\tilde{s} + 1).\)
    \end{enumerate}
    % For every \(\delta > 0\) there exists \(\lambda_0 \ge 1\) such that the following holds. For every integer \(\lambda \ge \lambda_0\), there exist constants \(\tilde{\varepsilon}, \tilde{s}\) depending only on \(\lambda\) such that \(\lambda = \lambda^{*, \OM}(\tilde{\eps} n, \tilde{s})\) and
    % \begin{align*}
    %     \Delta^{\OM}_{\ge 1}(\tilde \eps n, \lambda) = (4 \pm \delta) |\Delta^{\OM}_{\le -1}(\tilde \eps n, \lambda)| = (1 \pm \delta)/(\tilde{s} + 1).
    % \end{align*}
    Additionally\footnote{The subscript indicates dependency on $\lambda$, i.e., for all $c,C >0$ there exists $\lambda_0$ such that for all $\lambda \ge \lambda_0$ we have $\tilde \eps(\lambda) \le c/\lambda$ and $\tilde s(\lambda) \ge C$.} \(\tilde{\varepsilon}(\lambda) = o_{\lambda}( 1/\lambda )\) and \(\tilde{s}(\lambda) = \omega_{\lambda}(1)\). In particular, for every $\delta>0$, a sufficiently large \(\lambda_0\) guarantees that one may apply Lemmas~\ref{lem:pimp-relations} and \ref{lem:drift-bounds}.
\end{lemma}

\begin{proof}
    Take some arbitrary \(\delta \in (0, 1]\), \(\lambda_0\) large enough for Lemma~\ref{lem:drift-bounds}~\ref{lem:itm:drift-bounds-3}, and consider some \(\lambda \ge \lambda_0\). Since \(\lambda\) is a constant with respect to \(n\), we may also choose \(1/n \le \eps \le c\delta / \lambda\), with \(c\) the smallest constant appearing in Lemmas~\ref{lem:pimp-relations} and \ref{lem:drift-bounds}.

    With Lemma~\ref{lem:drift-bounds}~\ref{lem:itm:drift-bounds-3} we may estimate the negative contribution of the drift as 
    \begin{align*}
        \Delta^{\OM}_{\le -1} = -(1 \pm \delta) (1 - e^{-1})^\lambda. \numberthis \label{eq:lem:construction_equilibrium_OM_proof_1}
    \end{align*}

    For the positive contribution, recall that since we consider \(\OM\), the number of \(1\)-bits corresponds exactly to the fitness, i.e., we have \(\qimp^{\OM} = p^{\OM}_{\ge 1}\). As \(p^{\OM}_1 = \Delta^{\OM}_1\) by Definition~\ref{def:delta-p}, we may apply Lemma~\ref{lem:drift-bounds}~\ref{lem:itm:drift-bounds-2} for this pair \((\lambda, \eps)\) to find \(\Delta^{\OM}_{\ge 1} = (1 \pm \delta) p^{\OM}_{\ge 1} = (1 \pm \delta) \qimp^{\OM}\). Now using Lemma~\ref{lem:pimp-relations}~\ref{lem:itm:pimp-relations-1} gives \(\Delta^{\OM}_{\ge 1} = (1 \pm \delta)^2 \lambda \pimp^{\OM}\). 
    % \begin{align*}
    %     \Delta^{\OM}_{\ge 1} = (1 \pm \delta) p^{\OM}_{\ge 1} = (1 \pm \delta) \qimp^{\OM} = (1 \pm \delta)^2 \pimp^{\OM} \lambda.
    % \end{align*}
   
    Since \(\pimp^{\OM} = e^{-1} \eps \pm 10 \eps^2\) by Lemma~\ref{lem:pimp-relations}~\ref{lem:itm:pimp-relations-general}, and since we chose \(\eps \le c \delta / \lambda\) with \(\lambda \ge \lambda_0\) large enough, we may write \(\Delta^{\OM}_{\ge 1} = (1 \pm \delta)^3 e^{-1} \lambda \eps\). Now choosing\footnote{Note that this is indeed smaller than \(\delta c / \lambda\) as we assume \(\lambda_0\) to be large.} \(\eps := \tilde \eps = 4e (1 - e^{-1})^\lambda / \lambda\), we get
    \[
        \Delta^\OM_{\ge 1}(\tilde \eps n, \lambda) = (1 \pm \delta)^3 4(1 - e^{-1})^\lambda. \numberthis \label{eq:lem:construction_equilibrium_OM_proof_2}
    \]
    Equations \eqref{eq:lem:construction_equilibrium_OM_proof_1} and \eqref{eq:lem:construction_equilibrium_OM_proof_2} hold for every \(\delta\), so up to originally choosing a smaller \(\delta\), we have shown that \(\Delta^\OM_{\ge 1}(\tilde \eps n, \lambda) = (4 \pm \delta)|\Delta^\OM_{\le -1}(\tilde \eps n, \lambda)|\). 
    % If we choose \(\eps := \tilde \eps = 4e (1 - e^{-1})^\lambda / \lambda\), then we see from \eqref{eq:lem:construction_equilibrium_OM_proof_1} and \eqref{eq:lem:construction_equilibrium_OM_proof_2} then we get \todo{How about writing \(\subseteq \) instead of \(=\)?} \((1  \pm \delta)^3 \Delta^{\OM}_{\ge 1} = (1 \pm \delta) \cdot 4 |\Delta^{\OM}_{\le -1}|\), which is what we want (up to originally choosing a smaller \(\delta\) \todo{Rephrase a bit by saying: ``since this holds for arbitrary delta, it implies XXX''}). Note that this argument works because we assumed \(\lambda_0\) large enough, which implies \(\tilde \eps \le c \delta / \lambda\). Also, 
    Also, the fact that \(\tilde \eps = o_\lambda(1/\lambda)\) follows immediately from the formula.
    
    Given \(\lambda\) and \(\tilde \eps\) as above, the choice of \(\tilde s\) is constrained by the condition \(\lambda^*(\tilde \eps n, \tilde s) = \lambda\): it is the solution to \( 1/(\tilde s + 1) = \qimp^{\OM}(\tilde \eps n, \lambda)\). 
    % Since \(\Delta_{\ge 1} = (1 \pm \delta)\qimp^{\OM} \) the equality is proved (again, up to originally choosing a smaller \(\delta\)). 
    All that remains is to check that \(\tilde{s}\) has the correct order: the inequality above may be rewritten as \(\tilde s \; = 1/\qimp^{\OM} - 1 \) and we know from Lemma~\ref{lem:pimp-relations}~\ref{lem:itm:pimp-relations-1} that \(\qimp^{\OM} \le (1 + \delta) \lambda \pimp^{\OM} \le (1 + \delta) \lambda \tilde \eps\). In particular, this implies \(\tilde s \ge 1/((1 + \delta) \tilde \eps \lambda) - 1 = \omega_\lambda(1)\), since \(\tilde \eps = o(1/\lambda)\).
    % = \Theta( 1/(\lambda \tilde \varepsilon) ) = \omega_{\lambda}(1)\) using Lemma~\ref{lem:pimp-relations}, which concludes the proof.
    \qed

\end{proof}

The lemma above implies that for parameters \(\lambda, \tilde s\) and at distance \(\tilde \eps n\) from the optimum, the drift of \(Z\) for \onemax is roughly \(\Delta^\OM = \Delta^{\OM}_{\ge 1} + \Delta^{\OM}_{\le -1} \approx \tfrac{3}{4} \Delta^\OM_{\ge 1} > 0\).
Moreover, we want to show that the \saolea can not only pass this point, but continues all the way to the optimum. To this end, we define a more general potential function already used by~\cite{hevia2021arxiv} and~\cite{kaufmann2022arxiv}.

\begin{definition}
    \label{def:functions-h-g}
    We define \quad \(h(\lambda) := -\frac{1}{2} \log_F \lambda \) \quad and \quad \(g(x, \lambda) := \ZM(x) + h(\lambda)\).
    % \begin{align*}
    %     h(\lambda) := -\frac{1}{2} \log_F \lambda 
    %     \qquad \text{and} \qquad
    %     g(x, \lambda) := \ZM(x) + h(\lambda).
    % \end{align*}
\end{definition}

For convenience and following our notation \(Z^t = \ZM(x^t)\), we will write \(H^t = h(\lambda^t)\) and \(G^t = Z^t + H^t = g(x^t, \lambda^t)\).

\begin{lemma}
    \label{lem:drift_H}
    Let \(f \in \{ \OM, \DBV \}\). At all times \(t\) such that $\lambda^t \ge F$ we have 
    \begin{align*}
        \EE \left[ H^t - H^{t+1} \mid x^t , \lambda^t \right] = \tfrac{1}{2s}(1- (s+1)\qimp^{f}(x^t,\round{\lambda^t})).
    \end{align*}
    When \(\lambda^t < F\), we have \(\EE \left[ H^t - H^{t+1} \mid x^t , \lambda^t \right] \ge \tfrac{1}{2s}(1- (s+1)\qimp^{f}(x^t,\round{\lambda^t})).\)
\end{lemma}

\begin{proof}
    We have $H^{t+1} = -\frac{1}{2}\log_F(\lambda^t/F) = H^t+\frac{1}{2}$ in case of an improvement, and $H^{t+1} = H^t-1/2s$ otherwise. Hence, the drift of $H^t$ is
\begin{align*}
    \EE[H^t-H^{t+1} \mid x^t,\lambda^t] = -\frac{1}{2}\qimp^{\OM} + (1-\qimp^{f})\tfrac{1}{2s} = \tfrac{1}{2s}(1- (s+1)\qimp^{f}).\qquad
\end{align*}   

    The inequality in the case \(\lambda < F\) is simply obtained by recalling that \(\lambda^t\) may never drop below \(1\), which means that \(H^{t+1} \le H^t + 1/2\) in case of improvement.
   \qed
\end{proof}

We now show that for this choice of \(h\) and \(g\), the drift of $G^t$ for \OM is positive \emph{whenever} $Z^t \lesssim \tilde \eps n$ and $\lambda^t \ge \lambda-1$.

\begin{corollary}
    \label{cor:OM_positive_drift}
    Let $f=\OM$. There exists \(\lambda_0 \ge 1\) such that the following holds for all \(\lambda \ge \lambda_0\). Let $s = \tilde s = \tilde s(\lambda)$, $\eps = \tilde \eps = \tilde \eps(\lambda)$ be as in Lemma~\ref{lem:construction_equilibrium_OM}. There exist positive constants (with respect to \(n\)) \(\rho(\lambda)\) and \(\eps'(\lambda)\) such that if \(1 \le  Z^t \le (\varepsilon +\eps')n\) and \(\lambda^t \ge \lambda-1\), then 
    \begin{align*}
        \EE [ G^t - G^{t+1} \mid Z^t, \lambda^t ] \ge \rho.
    \end{align*}
\end{corollary}

\begin{proof}
    Let \(\delta > 0\) be a small constant and \(\lambda_0\) a constant large enough for Lemma~\ref{lem:construction_equilibrium_OM} to apply. Since \(\tilde \eps(\lambda)\) decays exponentially fast in \(\lambda\) by Lemma~\ref{lem:construction_equilibrium_OM}, we may assume that (up to possibly choosing \(\lambda_0\) larger, but still constant with respect to \(n\)) Lemmas~\ref{lem:pimp-relations} and \ref{lem:drift-bounds} hold for any \(\lambda \ge \lambda_0\) for \(\eps = \tilde \eps(\lambda)\).
    % \ref{lem:construction_equilibrium_OM} hold. 
    
    Let \(\lambda \ge \lambda_0\) be arbitrary and let \(\eps = \tilde \eps(\lambda), s = \tilde s(\lambda)\). By Lemma~\ref{lem:pimp-relations}~\ref{lem:itm:pimp-relations-2}, there exists a small \(\varepsilon' > 0\) such that \(\lambda^{*, \OM}((\varepsilon + \varepsilon')n, s) \ge \lambda - 1/4\).
    We seek to estimate the drift of \(G\) when \(Z^t \le (\varepsilon + \varepsilon') n\) and \(\lambda^t \ge \lambda - 1\). Using Lemma~\ref{lem:drift_H}, we see that this drift is  
    \begin{align*}
        \EE[ G^t - G^{t+1} \mid Z^t, \lambda^t ]
            &= \Delta^\OM_{\ge 1}(Z^t, \lambda^t) + \Delta^\OM_{\le -1}(Z^t, \lambda^t) \\
                &\qquad \qquad \qquad - \frac{1}{2}\qimp^{\OM}(Z^t, \lambda^t) + \frac{1}{2}(1 - \qimp^{\OM}(Z^t, \lambda^t)) / s.
    \end{align*}
    
    Since we consider \(f = \OM\), the expected number of bits gained \(\Delta_{\ge 1}^\OM(Z^t, \lambda^t)\) is larger than the probability \(\qimp^\OM(Z^t, \lambda^t)\) of finding an improvement. In particular, we may rewrite
    \begin{align*}
        \EE[ G^t - G^{t+1} \mid Z^t, \lambda^t ] \;
            &\ge \; \frac{1 - 1/s}{2} \qimp^{\OM}(Z^t, \lambda^t) + \frac{1}{2s} + \Delta_{\le -1}^\OM(Z^t, \lambda^t) \\
            &\ge \; \frac{1}{2s} + \Delta^\OM_{\le -1}(Z^t, \lambda^t), \numberthis \label{eq:drift-G-simpler-expression}
    \end{align*}
    as a choice of \(\lambda_0\) large enough guarantees that \(s > 1\).
    To evaluate this expression, we now bound \(\Delta_{\le -1}^\OM(Z^t, \lambda^t)\). From Lemma~\ref{lem:drift-bounds}~\ref{lem:itm:drift-bounds-3} we know that \((1 - \delta) (1 - e^{-1})^\lambda \le | \Delta^\OM_{\le -1}(\eps n, \lambda) | \), and from Lemma~\ref{lem:construction_equilibrium_OM} we known that \(|\Delta^\OM_{\le -1}(\eps n, \lambda ) | \le \frac{1}{4 - \delta} \Delta^\OM_{\ge 1}(\eps n, \lambda)\); together they imply
    \begin{align*}
        (1 - e^{-1})^\lambda \le \frac{1}{(1 - \delta)(4 - \delta)} \Delta^\OM_{\ge 1}(\eps n, \lambda). \numberthis \label{eq:OM_positive_drift_proof_1}
    \end{align*} 
    By Lemma~\ref{lem:drift-bounds}~\ref{lem:itm:drift-bounds-2} we have \(\Delta^\OM_{\ge 1}(\eps n, \lambda) \le (1 + \delta) \Delta^\OM_1(\eps n, \lambda)\), and as we are considering \(f = \OM\), we have \(\Delta^\OM_1(\eps n, \lambda) = p^\OM_1 (\eps n, \lambda) \le \qimp^\OM(\eps n, \lambda)\)). By construction, \(\eps = \tilde \eps\) and \(s = \tilde s\) are chosen so that \(\lambda^{*, \OM}(\eps n, s) = \lambda\), i.e., \(\qimp^\OM(\eps n, \lambda) = 1/(1 + s)\) so together with~\eqref{eq:OM_positive_drift_proof_1} this gives 
    \begin{align*}
        (1 - e^{-1})^\lambda \le \frac{1 + \delta}{(1 - \delta) (4 - \delta)} \cdot  \frac{1}{1 + s}.
    \end{align*}
    To estimate \(\Delta_{\le -1}^\OM(Z^t, \lambda^t)\), we first observe that \(\Delta_{\le -1}^\OM(Z^t, \cdot)\) is decreasing\footnote{Indeed, we may express \(|\Delta^\OM_{\le -1}( Z, \lambda )| = \sum_{i=1}^\infty{ p^\OM_{\le - i}(Z, \lambda) } = \sum_{i=1}^\infty{ (p^\OM_{\le - i}(Z, 1))^\lambda }\), which is decreasing with \(\lambda\).}. Hence, applying Lemma~\ref{lem:drift-bounds}~\ref{lem:itm:drift-bounds-3}, now for worst case \(\lambda^t = \lambda-1\), we have \(|\Delta^\OM_{\le -1}(Z^t, \lambda^t)| \le (1 + \delta) (1 - e^{-1})^{\lambda - 1} \), which by the above must be at most 
    \begin{align*}
        |\Delta^\OM_{\le -1}(Z^t, \lambda^t)| \; \le \; \frac{(1 + \delta)^2}{(4 - \delta)(1 - \delta)} \cdot \frac{1}{(1 - e^{-1})(1 + s)} \; \le \; \frac{1}{2(1 + s)},
    \end{align*}
    with the last step holding for a choice of \(\delta\) small enough. Replacing in~\eqref{eq:drift-G-simpler-expression} gives 
    \begin{align*}
        \EE[ G^t - G^{t+1} \mid Z^t, \lambda^t ]
            &\ge \frac{1}{2s} - \frac{1}{2(1 + s)} 
            = \frac{1}{2 s(1 + s)}.
    \end{align*}
    Setting \(\rho(\lambda) = 1 / (2 \tilde s (1 + \tilde s))\), a constant with respect to \(n\), concludes the proof.

    \qed

\end{proof}

We have just shown that when within distance at most \(\tilde \varepsilon n\) from the optimum, \onemax has drift \emph{towards} the optimum. We now turn to \dynBV: the following lemma states that at distance \(\tilde \varepsilon n\) from the optimum, \dynBV has drift \emph{away} from the optimum. 

\begin{lemma}
    \label{lem:construction_inefficient_DBV}
    Let \(f = \DBV\). There exists \(\lambda_0 \ge 1\) such that for all \(\lambda \ge \lambda_0\) there are \(\nu,\eps' > 0\), constants with respect to \(n\), such that the following holds. Let \(\tilde s = \tilde{s}(\lambda), \tilde \varepsilon = \tilde{\varepsilon}(\lambda)\) as in Lemma~\ref{lem:construction_equilibrium_OM}. If \(Z^t = (\tilde \varepsilon \pm \varepsilon') n\) and \(\lambda^t \le \lambda^{*,\DBV}(\tilde \varepsilon n, \tilde{s}) + 1\) we have 
    \begin{align*}
        \EE[Z^t - Z^{t+1} \mid x^t, \lambda^t] \le -\nu.
    \end{align*}
\end{lemma}

\begin{proof}
    Let \(\delta\) be a small constant and choose a large \(\lambda_0\). Consider some \(\lambda \ge \lambda_0\) and let \(\tilde \eps = \tilde \eps (\lambda), \tilde s = \tilde s (\lambda)\). Since \(\tilde \eps = o(1/\lambda)\) by Lemma~\ref{lem:construction_equilibrium_OM}, we may assume \(\tilde \eps\) is small enough to apply first Lemma~\ref{lem:pimp-relations}~\ref{lem:itm:pimp-relations-2} and then Lemma~\ref{lem:pimp-relations}~\ref{lem:itm:pimp-relations-3}: there exists an \(\eps'\) such that 
    \[
       \lambda^{*, \DBV}((\tilde \eps + \eps')n, \tilde s) \; \ge \; \lambda^{*, \DBV}(\tilde \eps n, \tilde s) - 1/4 \; \ge \; 0.5 \lambda^\OM(\tilde \eps, \tilde s) \; \ge \; 0.4 \lambda.
    \]
    
    Up to taking a larger \(\lambda_0\) (giving a smaller \(\tilde \eps\)) and a smaller \(\eps '\), we may assume that \(\tilde \eps + \eps' < \tfrac{0.4c \delta}{\lambda}\), with \(c\) the smallest constant appearing in Lemmas~\ref{lem:pimp-relations} and  \ref{lem:drift-bounds}. In particular, we may apply those lemmas with \(\tilde \eps + \eps'\) and \(\lambda^{*, f}((\tilde \eps + \eps') n, \tilde s)\) for both \(f = \OM\) and \(f = \DBV\).

    % Combining both displays above gives 
    % \begin{align}
    %     \lambda^{*, \DBV}((\tilde \eps n, \tilde s))
    %         \ge \lambda^{*, \DBV}((\tilde \eps n, \tilde s))
    % \end{align}

    % \(\), since \(\lambda \ge \lambda_0\) is assumed large enough. The same lemma guarantees that \(0.6\lambda^{*, \OM}(\tilde \eps n, \tilde s) \le \lambda^{*, \DBV}(\tilde \eps n, \tilde s)\), so we may also apply Lemma~\ref{lem:drift-bounds} with \(\lambda^{*, \DBV}(\tilde \eps n, \tilde s)\) and \(\tilde{\eps} + \eps'\) for the chosen \(\delta\).

    % For this choice of \(Z^t, \lambda^t\), the drift of \(Z\) for \DBV is 
    Using Lemma~\ref{lem:drift-bounds}~\ref{lem:itm:drift-bounds-1}, the drift we seek to bound may be expressed as
    \begin{align*}
        \Delta^{\DBV}_{\ge 1}(Z^t, \lambda^t) + \Delta^{\DBV}_{\le -1}(Z^t, \lambda^t) \le \Delta^{\OM}_{\ge 1}(Z^t, \lambda^t) + \Delta^{\OM}_{\le -1}(Z^t, \lambda^t),
    \end{align*}
    and we aim at showing that this is negative. Lemma~\ref{lem:drift-bounds}~\ref{lem:itm:drift-bounds-2} guarantees that the positive contribution of this drift is at most 
    \begin{align*}
        \Delta^\OM_{\ge 1}(Z^t, \lambda^t)
            &\le (1 + \delta) \Delta^{\OM}_1(Z^t, \lambda^t)
            \le (1 + \delta) \qimp^{\OM}(Z^t, \lambda^t) \\ 
            &\le (1 + \delta) \qimp^\OM \left((\tilde \eps + \eps')n , \lambda^{*, \DBV}((\tilde \eps + \eps') n, \tilde s) + 2 \right), %\\
            % &\le (1 + \delta) \qimp^\OM((\tilde \eps + \eps')n , \lambda^{*, \OM}((\tilde \eps + \eps') n, \tilde s)) \\
            % &\le (1+\delta)/(\tilde s + 1),
    \end{align*}
    where the last step follows from the monotonicity of \(\qimp^{\OM}\) in both \(Z\) and \(\lambda\). We know from Lemma~\ref{lem:pimp-relations}~\ref{lem:itm:pimp-relations-3} that \(\lambda^{*, \DBV}((\tilde \eps + \eps') n, \tilde s) + 2 \le 0.6 \lambda^{*, \OM}((\tilde \eps + \eps') n, \tilde s) + 2 \le \lambda^{*, \OM}((\tilde \eps + \eps') n, \tilde s)\) as we assume \(\lambda_0\) large enough.
     
    In particular, still using the monotonicity of \(\qimp^\OM\), this implies, by the definition of \(\lambda^{*, \OM}\), that the positive contribution to the drift is at most 
    \[
        (1 + \delta) \qimp^\OM\left((\tilde \eps + \eps')n , \lambda^{*, \OM}((\tilde \eps + \eps') n, \tilde s)\right) = (1 + \delta)/(\tilde s + 1).
    \]
    
    Let us now show that the negative contribution is larger. Recalling that \(\lambda^t \le \lambda^{*, \DBV}(\tilde \varepsilon n, \tilde s) + 1\) by assumption, an application of Lemma~\ref{lem:pimp-relations}~\ref{lem:itm:pimp-relations-3} then gives \(\lambda^t \le \lambda^{*, \DBV}(\tilde \varepsilon n, \tilde s) + 1 \le 0.7\lambda\). 
    Now applying Lemma~\ref{lem:drift-bounds}~\ref{lem:itm:drift-bounds-3} we find
    \begin{align*}
         | \Delta_{\le -1}^{\OM}(Z^t, \lambda^t ) |
            &\ge (1 - \delta) (1 - e^{-1})^{ 0.7 \lambda}. \numberthis \label{eq:lem:construction_inefficient_DBV_proof_1}
    \end{align*}    
    By Lemma~\ref{lem:construction_equilibrium_OM} we have \(|\Delta_{\le -1}^{\OM}( \tilde \varepsilon n, \lambda)| \ge \frac{1}{4 + \delta} \Delta^{\OM}_{\ge 1}\); as for \OM we have \(\Delta^\OM_{\ge 1} \ge \qimp^\OM\), we deduce \(|\Delta^{\OM}_{\le -1}(\tilde \eps n, \lambda)| \ge \frac{1}{4 + \delta} \qimp^\OM(\tilde \eps n, \lambda) = \frac{1}{(4 + \delta)(1 + \tilde s)}\) by definition of \(\tilde \eps\) and \(\tilde s\). An application of Lemma~\ref{lem:drift-bounds}~\ref{lem:itm:drift-bounds-3} also gives \(|\Delta^\OM_{\le -1}(\tilde \eps n, \lambda)| \le (1 + \delta) (1 - e^{-1})^\lambda\), which implies 
    \begin{align*}
        (1 + \delta)(1 - e^{-1})^\lambda \ge \frac{1}{(4 + \delta)(1 + \tilde s)}. \numberthis \label{eq:lem:construction_inefficient_DBV_proof_2}
    \end{align*}
    Combining~\eqref{eq:lem:construction_inefficient_DBV_proof_1} and \eqref{eq:lem:construction_inefficient_DBV_proof_2} gives
    % By Lemma~\ref{lem:construction_equilibrium_OM}, we have \(|\Delta_{\le -1}^{\OM}( \tilde \varepsilon n, \lambda)| \ge (1 - \delta) \frac{1}{4(\tilde s+1)}\), and a new application of Lemma~\ref{lem:drift-bounds}~\ref{lem:itm:drift-bounds-3} gives \((1 - \delta) / (4(\tilde s+1)) \le (1 + \delta) (1 - e^{-1})^\lambda \).\\
    % Replacing in the display above gives 
    % \begin{align*}
    %      |\Delta_{\le -1}^{\OM}(Z^t, \lambda^t) |
    %         &\ge \frac{(1 - \delta)^2}{1 + \delta} (1 - e^{-1})^{-0.3 \lambda} \frac{1}{4(\tilde s+1)} \ge \frac{2}{\tilde s+1},
    % \end{align*}
    \begin{align*}
        | \Delta_{\le -1}^{\OM}(Z^t, \lambda^t) |
            &\ge \frac{(1 - \delta) \cdot (1 - e^{-1})^{-0.3 \lambda} }{(1 + \delta)(4 + \delta)} \cdot \frac{1}{1 + \tilde s} \ge \frac{2}{1 + \tilde s},
    \end{align*}
    as \(\lambda\) may be chosen large enough compared to \(\delta\).
    In the range of \(Z^t, \lambda^t\) from the lemma, we see that 
    \begin{align*}
        \EE[ Z^t - Z^{t+1} \mid x^t, \lambda^t ]
            &\le (1 + \delta) \frac{1}{\tilde s+1} - \frac{2}{\tilde s+1}
            \le -\frac{1}{2(\tilde s+1)},
    \end{align*}
    which concludes the proof.\qed
\end{proof}

Corollary~\ref{cor:OM_positive_drift} and Lemma~\ref{lem:construction_inefficient_DBV} respectively give drift towards the optimum for \onemax and away from the optimum for \dynBV. We are almost ready to state and prove our final theorem. A last step before this is the following lemma saying \(\lambda \le n^2\) at all steps and that \(Z\) never changes by more than \(\log n\).

\begin{lemma}\label{lem:small-steps}
Consider the \saolea on \onemax or \dynBV. There exists \(T = n^{\omega(1)}\) such that, with probability $1-n^{-\omega(1)}$, either the optimum is found within the first \(T\) steps, or $\lambda^t \le n^2$ and $|Z^t-Z^{t+1}| \le \log n$ both hold during those \(T\) steps.
\end{lemma}
\begin{proof}
In order to grow above $n^2$, there would need to be a non-improving step when $\lambda^t \ge n^2/F^{1/s}$. Let us call this event $\mathcal E^t$. The probability that a fixed offspring finds an improvement is at least $1/n \cdot (1-1/n)^{n-1} \geq 1/(en)$ (using Lemma~\ref{lem:basic}~\ref{lem:itm:basic-1} with \(y = n\)), even if the algorithm is just one step away from the optimum. Hence, the probability that none of $\lambda^t$ offspring finds an improvement is at most 
\begin{align*}
\Pr[\mathcal E^t] \le  (1-\tfrac{1}{en})^{\lambda^t} \le e^{- \lambda^t /(en)} = e^{-\Omega(n)}.
\end{align*}
using Lemma \ref{lem:basic}~\ref{lem:itm:basic-1} with \(y = e n\) for the second inequality. Even a union bound over a superpolynomial number of steps leaves the error probability $e^{-\Omega(n)}$ unchanged. This proves the first statement.

For the step sizes, condition on $\bar{ \mathcal E}^t$. We can bound the difference $|Z^t-Z^{t+1}|$ from above by the maximal number of bits that any offspring flips. The probability that a fixed offspring flips exactly $k \ge 1$ bits is at most 
\begin{align*}
\binom{n}{k} \cdot \Big(\frac{1}{n}\Big)^k \cdot (1-1/n)^{n-k} \; \le \; \frac{1}{k!},
\end{align*}
and the probability that it flips at least $k$ bits is at most $\sum_{i\ge k} 1/i! \le 2/k!$, since the factorials are bounded by a geometric sum as $1/(k+1)! \le 1/(2\cdot k!)$. For $k = \log n$ this bound is $2/k! = e^{-\Omega(k\log k)} = e^{-\omega(\log n)} = n^{-\omega(1)}$, and the same holds after a union bound over at most $n^2$ offspring, and after a union bound over a superpolynomial (suitably chosen) number of steps.\qed
\end{proof}

Now we are ready to formulate and prove the main result of this paper.

\begin{theorem}
    \label{thm:existence_threshold}
    Let $F=1+\eta$ for some $\eta = \omega(\log n/n) \cap o(1/\log n)$. There exist constants \(\eps, s\) such that with high probability the \saolea with success rate~$s$, update strength $F$ and mutation rate $1/n$ starting with $Z^0 =\eps n$,
    \begin{enumerate}[(a)]
        \item finds the optimum of \onemax in $O(n)$ generations;\label{thm:itm:existence_threshold_1}
        \item does not find the optimum of \dynBV in a polynomial number of generations.\label{thm:itm:existence_threshold_2}
        \end{enumerate}
\end{theorem}

\begin{proof}
    Let us first recall the intuition presented in Section~\ref{sec:sketch} and sketch how we intend to prove the theorem. The cornerstone of our argument is Lemma~\ref{lem:construction_equilibrium_OM}. It states, that given a large enough (but constant) \(\lambda\), we may find \(\tilde \eps\) and \(\tilde s\) such that 1) \(\lambda^{*, \OM}(\tilde \eps n, \tilde s) = \lambda\) and 2) the drift of \(Z\) for \OM is \emph{mildly} dominated by the positive term \(\Delta^\OM_{\ge 1}\). We choose \(s = \tilde s\) and choose to start the optimisation (for both \OM and \DBV) exactly with \(\tilde \eps n\) zero-bits. For \OM, we then argue that \(\lambda^t\) quickly gets, and remains close to \(\lambda^{*, \OM}(\tilde \eps n, \tilde s)\) and then use Corollary~\ref{cor:OM_positive_drift} to conclude that the process has positive drift towards the optimum, and that optimisation must be efficient. For this same choice of \(s\) and starting position \(\tilde \eps n\), the negative part of the drift for \DBV is much higher and dominates the positive contribution. We have proved this in Lemma~\ref{lem:construction_inefficient_DBV} and this implies that \DBV takes a superpolynomial number of steps before finding the optimum.

    We now move to the formal proof. Let \(\delta\) be a sufficiently small constant and choose \(\lambda\) large. Let \(s := \tilde{s}(\lambda)\) and \(\varepsilon = \tilde{\varepsilon}(\lambda)\) as in Lemma~\ref{lem:construction_equilibrium_OM}. 
    Since \(\lambda^{*, \DBV}(\eps n, s) \ge 0.5 \lambda^{*,\OM}(\eps n, s) = 0.5\lambda\), we may assume that Lemma~\ref{lem:drift-bounds} holds for \(\lambda^{*, \DBV}(\varepsilon n, s)\).

    % We first prove~\ref{thm:itm:existence_threshold_1}. 
    \proofitem{Proof of~\ref{thm:itm:existence_threshold_1}}
    Set $f:=\onemax$, and let $\eps' = \eps'(\lambda)$ be small enough (but independent of \(n\)) to apply Corollary~\ref{cor:OM_positive_drift}.
    We will first argue that $\lambda^t$ quickly reaches a value of at least $\lambda - 7/8$ and stays above $\lambda - 1$ afterwards. First, we observe that if \(\lambda^t \le \lambda - 7/8\), then \(\round{\lambda^t} \le \lambda^t + 1/2 \le \lambda - 3/8\). By Lemma~\ref{lem:drift_H}, the drift of \(H\) is at least
    \begin{align*}
        E[H^t-H^{t+1} \mid Z^t,\lambda^t] 
            &\ge \tfrac{1}{2s}(1-(s+1)\qimp^{\OM}(Z^t,\round{\lambda^t})).
    \end{align*}
    For \OM, the probability to find an improvement \(\qimp^{\OM}\) --- that is, to increase the number of \(1\)-bits --- is increasing in both the number of \(0\)-bits \(Z\) and the number of children \(\lambda\) generated. In particular, at each step when \(Z^t \le (\eps + \eps')n\) and \(\lambda^t \le \lambda - 7/8\), we must have 
    \begin{align*}
        E[H^t-H^{t+1} \mid Z^t,\lambda^t]
            &\ge \frac{1}{2s} (1 - (s+1)\qimp^{\OM}((\eps + \eps')n, \lambda - 3/8)). \numberthis \label{eq:thm:existence_threshold_1} 
    \end{align*}
    To prove that this drift is a positive constant, we need to understand the term \(\qimp^{\OM}((\eps + \eps')n, \lambda - 3/8)\). Recall that for all \(\bar \lambda\) we have \(\qimp^{\OM}(Z, \bar \lambda) = 1 - (1 - \pimp^{\OM}(Z))^{\bar \lambda}\) by Definition~\ref{def:lambdastar} so that
    \begin{align*}
        \qimp^{\OM}( (\eps + \eps') n, \lambda - 3/8) 
            \; = \; \qimp^{\OM}( (\eps + \eps') n, \lambda^{*, \OM}((\eps + \eps')n, s)) - d_1 \; = \; \frac{1}{1 + s} - d_1 \numberthis \label{eq:thm:existence_threshold_2}
    \end{align*}
    with \(d_1 = (1 - \pimp^{\OM}((\eps + \eps')n))^{\lambda - 3/8} - (1 - \pimp^{\OM}((\eps + \eps')n))^{\lambda^{*, \OM}((\eps + \eps')n, s))}\). Here the second inequality holds since \(\qimp^{\OM}( (\eps + \eps') n, \lambda^{*, \OM}((\eps + \eps')n, s)) = \tfrac{1}{1 + s}\) by definition.

    To estimate \(d_1\), we apply Lemma~\ref{lem:pimp-relations}~\ref{lem:itm:pimp-relations-2} to get $\lambda^{*, \OM}(( \eps+\eps')n, s) \ge \lambda - 1/4 = \lambda - 3/8 + 1/8$. This then implies that 
    \begin{align*}
        d_1 
            &\ge (1 - \pimp^{\OM}((\eps + \eps')n)^{\lambda - 3/8} \cdot (1 - (1 - \pimp^{\OM}((\eps + \eps')n))^{1/8}),
    \end{align*}
    is a positive constant since \(\lambda\), \(\eps\) and \(\eps'\) are constants independent of \(n\), and \(\pimp^{\OM}((\eps + \eps') n)\) is of order \(\eps + \eps'\) by Lemma~\ref{lem:pimp-relations}~\ref{lem:itm:pimp-relations-general}. Replacing in \eqref{eq:thm:existence_threshold_2} and in \eqref{eq:thm:existence_threshold_1} gives a positive drift
    \begin{align*}
        E[H^t-H^{t+1} \mid Z^t,\lambda^t]
            &\ge \frac{(s+1) d_1}{2s} =: d.
    \end{align*}

    % Recall that \(\qimp = 1 - (1 - \pimp)^\lambda \) by Definition~\ref{def:lambdastar}, and note that by definition of $\eps = \tilde \eps$, we have $\qimp( \eps n,\lambda) = 1/(s+1)$. This implies
    % \begin{align*}
    %     \qimp( \eps n,\lambda - 1/4) 
    %         = \qimp( \eps n,\lambda) + (1 - \pimp(\eps n))^{\lambda} - (1 - \pimp(\eps n))^{\lambda - 1/4} 
    %         = \tfrac{1}{s+1} - d_1.
    % \end{align*}
    % Here, \(d_1 = (1 - \pimp(\eps n))^{\lambda - 1/4} - (1 - \pimp(\eps n))^{\lambda}\) is a small positive constant since \(\lambda, \eps\) are constant independent of \(n\), and \(\pimp(\eps n)\) is of order \(\eps\) by Lemma~\ref{lem:pimp-relations}.
    % Up to potentially reducing \(\varepsilon'\), we may apply Lemma~\ref{lem:pimp-relations}~\ref{lem:itm:pimp-relations-2} and get $\lambda^*(( \eps+\eps')n, s) \ge \lambda - 1/4$, or reformulated
    % \begin{align*}
    %     \qimp(( \eps+\eps') n,\lambda-1/4) \le \tfrac{1}{s+1} - d_1.
    % \end{align*}
    % Therefore, $\round{\lambda - 7/8} \le \lambda - 3/8 < \lambda - 1/4 \le \lambda^*(( \eps+\eps')n, s)$, and by Lemma~\ref{lem:drift_H} we have for all $1<Z^t \le ( \eps + \eps')n$ and all $\lambda^t \le \lambda - 7/8$,
    % \begin{align*}
    %     E[H^t-H^{t+1} \mid Z^t,\lambda^t] & =   \tfrac{1}{2s}(1-(s+1)\qimp(Z^t,\round{\lambda^t})) \\
    %     & \ge  \tfrac{1}{2s}(1-(s+1)\qimp((\eps+\eps')n,\lambda - 3/8)) \ge \frac{d_1 (s+1)}{2s} =: d. 
    % \end{align*}

    Therefore, $H^t$ has positive (constant) drift as long as $1<Z^t \le ( \eps + \eps')n$ and $\lambda^t \le \lambda - 7/8$. By Lemma~\ref{lem:small-steps}, we may assume that $Z^t$ changes by at most $\log n$ in every of the first $\eps'n/(2\log n)$ steps. In particular, we have $Z^t \le (\eps +\eps'/2)n$ during this time. Hence, $H^t$ has constant positive drift until either $\eps'n/(2\log n)$ steps are over or until it hits $-\frac{1}{2} \log_F(\lambda-7/8)$. Since $H^0 = -\frac{1}{2}\log_F(1) = 0$, by the additive drift theorem, the latter option happens after at most $\tfrac{1}{2d}\log_F (\lambda - 7/8) = O(1/ \log F) = O(1/\eta)$ steps in expectation. Since \(\eta = \omega(\log n / n)\) by assumption, this is at most \(o(n / \log n)\). Moreover, the same holds with high probability after at most $2 \tfrac{1}{2d}\log_F (\lambda-7/8) = O(1/\eta) = o(n / \log n)$ steps by concentration of hitting times for additive drift, Theorem~\ref{thm:AdditiveDrift}, since $H$ changes by definition by at most one in each step and \(2 \tfrac{1}{2d}\log_F (\lambda-7/8) = \Omega(1/\eta) = \omega(\log n)\). %Since $1/\eta = o(n/\log n)$, 
    Hence, with high probability, $\lambda^t \ge \lambda - 7/8$ happens before $Z^t$ reaches $( \eps+\eps'/2)n$. Let us call the first such point in time $t_0$.

    Now we show that after time $t_0$, the population size $\lambda^t$ stays above $\lambda - 1$ and \(Z^t\) remains below \((\eps + \eps') n\). To this intent, we wish to apply the Negative Drift Theorem~\ref{thm:NegativeDrift} to \(H^t\): for $\lambda^t$ to decrease from $\lambda-7/8$ to $\lambda-1$ (while \(Z^t \le (\eps + \eps')n\)) the potential must increase by $\ell := \frac{1}{2}\log_F(\lambda - 7/8) - \frac{1}{2}\log_F(\lambda - 1) = \Omega(1/\log F) = \Omega(1/\eta) = \omega(\log n)$.
    We have already established that \(H^t\) has at least positive constant drift whenever \(\lambda^t \le \lambda - 7/8\) and \(Z^t \le (\eps + \eps')n\), so condition~\ref{thm:itm:NegativeDrift-1} of the theorem holds. As \(|H^{t+1} - H^t|\) is bounded by \(1/2\), the second condition~\ref{thm:itm:NegativeDrift-2} also trivially holds for a constant \(r\) large enough. This choice of \(r = O(1)\) guarantees that condition~\ref{thm:itm:NegativeDrift-3} holds since \(\ell = \omega(\log n)\).
    By the Negative Drift Theorem (Theorem~\ref{thm:NegativeDrift}), we then conclude that with high probability this does not happen within $n^2$ steps. 

    We have shown that \(\lambda^t\) is very unlikely to drop below \(\lambda -1\) while \(Z^t \le (\eps + \eps')n\), and now prove that \(Z^t\) is also unlikely to exceed \((\eps + \eps')n\) while \(\lambda^t \ge \lambda - 1\).
    By definition of \(G^t = Z^t + H^t = Z^t - \frac{1}{2} \log_F(\lambda^t)\), we must have \(Z^t \ge G^t \ge Z^t - \frac{\log \lambda^t}{2 \log F} \ge Z^t - o( n )\), since we may assume by Lemma~\ref{lem:small-steps} that \(\lambda^t \le n^2\) for a superpolynomial number of steps, and since \(1/\log F = O(1/\eta) = o(n / \log n)\). In particular, to have \(Z^t \ge (\eps + \eps')n\), we must have \(G^t \ge (\eps + \eps')n - o(n) \ge (\eps + \tfrac{3}{4}\eps')n\), as \(\eps'\) is a constant (possibly small, but independent of \(n\)). As we start with a value \(G^{t_0} \le Z^{t_0} \le (\eps + \tfrac{1}{2}\eps') n\), \(Z^t\) may only exceed \((\eps + \eps')n\) if \(G\) increases by \(\ell := \tfrac{1}{4}\eps' n = \Omega(n)\). By Corollary~\ref{cor:OM_positive_drift} we know that \(G^t\) has constant positive drift while \(Z^t \le (\eps + \eps')n\) and \(\lambda^t \ge \lambda - 1\), i.e., condition~\ref{thm:itm:NegativeDrift-1} of Theorem~\ref{thm:NegativeDrift} holds. By Lemma~\ref{lem:small-steps}, we have \(|Z^t - Z^{t+1}| \le \log n\) with high probability for a superpolynomial number of steps. During those steps, we hence have \(|G^t - G^{t+1}| \le |Z^t - Z^{t+1}| + |H^t - H^{t+1}| \le \log n + 1/2\), so condition~\ref{thm:itm:NegativeDrift-2} also holds for \(r = O(\log n)\). The last condition~\ref{thm:itm:NegativeDrift-3} trivially holds for our choice of \(\ell = \Omega(n)\) and \(r = O(\log n)\).
    Hence, by the negative drift Theorem~\ref{thm:NegativeDrift}, the probability that \(G^t\) drops by \(\ell\) within \(n^{2}\) steps is at most \(e^{- \Omega( \frac{n}{\log^2 n} ) } = o(1)\).

    Summing up, we have established that with high probability,
    \begin{enumerate}[(i)]
        \item the process reaches a state with \(\lambda^t \le \lambda - 7/8\) and \(Z^t \le (\eps + \tfrac{1}{2}\eps')n\) within \(t_0 = o(n)\) steps;
        \item after that time, we have \(\lambda^t \ge \lambda - 1 \) and \(Z^t \le (\eps + \eps')n\) for at least $n^2$ steps; 
    \end{enumerate}
    and by Lemma~\ref{lem:small-steps} we may assume that \(\lambda^t \le n^2\) and \(|Z^t - Z^{t+1}|\le \log n\) during all those \(t_0 + n^2\) steps.
    Using those assumptions, we now wish to show that with very high probability \(Z^t = 0\) is attained during those \(n^2\) steps following time \(t_0\). 
    
    We know that the potential \(G\) starts from a value \(G^{t_0} = Z^{t_0} + H^{t_0} \le Z^{t_0}\). The upper bound on \(\lambda\) also guarantees that \(G^t \ge - \tfrac{1}{2} \log_F n^2\), with equality holding only when \(Z^t = 0\). In particular, if \(G^t\) drops by at least \(Z^{t_0} + \tfrac{1}{2} \log_F n^2 = O(n) + O(\tfrac{n^2}{\eta}) = O(n)\), then \(Z^t = 0\) must happen. 
    We know from Corollary~\ref{cor:OM_positive_drift} that the drift of $G^t$ is a positive constant as long as $Z^t > 0$, and since \(|G^t - G^{t+1}| \le |Z^t - Z^{t+1}| + |H^t - H^{t+1}| \le \log n + 1/2\), we may apply the Additive Drift Theorem~\ref{thm:AdditiveDrift}: the probability of not having hit \(Z^t = 0\) in \(2n / \rho\) steps is at most \(\exp \left( - \Omega\left( \frac{n}{\log^2 n} \right)\right) = o(1)\). This proves \ref{thm:itm:existence_threshold_1}.

    % Since we assume \(\lambda^t \le n^2\), we must also have \(G^t \ge - \tfrac{1}{2} \log_F n^2 = - O(n)\), with equality holding only when \(Z^t = 0\). In particular, as we start from \(G^{t_0} = Z^{t_0} + H^{t_0} \le Z^{t_0} = O(n)\), if \(G^t\) decreases by at least \(Z_{t_0} + \)with high probability the event $Z^t =0$ happens within $O(n)$ steps by the tail bounds on hitting times for additive drift, Theorem~\ref{thm:AdditiveDrift}. This proves \ref{thm:itm:existence_threshold_1}.  

    % By Corollary~\ref{cor:OM_positive_drift} the potential $G^t$ has a positive drift during those $n^2$ steps, or until $Z^t$ hits zero. Since we assume \(\lambda^t \le n^2\), we must have \(G^t \ge - \tfrac{1}{2} \log_F n^2 = - O(n)\), with equality holding only when \(Z^t = 0\).
    % Since we start with a potential \(G^{t_0}\) of at most $Z^{t_0} + H^{t_0} \le Z^{t_0} = O(n)$, with high probability the event $Z^t =0$ happens within $O(n)$ steps by the tail bounds on hitting times for additive drift, Theorem~\ref{thm:AdditiveDrift}. This proves \ref{thm:itm:existence_threshold_1}.  
    
    \proofitem{Proof of~\ref{thm:itm:existence_threshold_2}} 
    We now set $f:= \DBV$, and we work directly with the potential $Z^t$ instead of $G^t$. As in the previous case above, \(\lambda^t\) and \(Z^t\) influence each other, so to be able to conclude we will show that \(\lambda^t\) is unlikely to go above \(\lambda^t \le \lambda^{*,\DBV}(\varepsilon n,  s) + 1\) while \(Z^t \ge (\eps - \eps')n\), and that conversely we should expect \(Z^t \ge (\eps - \eps')n\) as long as \(\lambda^t \le \lambda^{*,\DBV}(\varepsilon n,  s) + 1\).
    
    Let $\eps' >0$ be the constant from Lemma~\ref{lem:construction_inefficient_DBV}. Assume that \(Z^t \ge (\eps - \eps') n\) and that \(\lambda^t \ge \lambda^{*,\DBV}(\varepsilon n,  s) + 7/8\). From Lemma~\ref{lem:drift_H}, we know that the drift of \(H^t\) is 
    \begin{align*}
        \EE[H^t - H^{t+1} \mid Z^t, \lambda^t] 
            &= \frac{1}{2s}(1 - (s + 1)\qimp^\DBV(Z^t, \round{\lambda^t})) \\
            &\le \frac{1}{2s}(1 - (s+1) \qimp^\DBV((\eps - \eps') n, \lambda^{*,\DBV}(\varepsilon n,  s) + 3/8)).
    \end{align*}
    Here the last step is derived as in the proof of~\ref{thm:itm:existence_threshold_1}: \(\qimp^\DBV\) is increasing in both \(Z^t\) and \(\lambda^t\), and \(\lambda^t \ge \lambda^{*,\DBV}(\varepsilon n,  s) + 7/8\) implies \(\round{\lambda^t} \ge \lambda^{*,\DBV}(\varepsilon n,  s) + 3/8 \). As above, we may write 
    \begin{align*}
        &\qimp^\DBV((\eps - \eps') n, \lambda^{*, \DBV}(\eps n, s) + 3/8) \\
            & \hspace{4cm}= \qimp^\DBV((\eps - \eps') n, \lambda^{*, \DBV}((\eps - \eps')n, s)) + d_2 = \frac{1}{1 + s} + d_2,
    \end{align*}
    where \(d_2 = (1 - \pimp^{\DBV}((\eps - \eps')n))^{\lambda^{*, \DBV}((\eps - \eps')n, s)} - (1 - \pimp^{\DBV}((\eps - \eps')n))^{\lambda^{*,\DBV}(\varepsilon n,  s) + 3/8}\) is a (possibly small, but independent of \(n\)) constant since \(\lambda^{*, \DBV}((\varepsilon - \varepsilon')n, s) \le \lambda^{*, \DBV}(\eps n , s) + 1/4 < \lambda^{*, \DBV}(\eps n , s) + 3/8\) by Lemma~\ref{lem:pimp-relations}~\ref{lem:itm:pimp-relations-2}. Replacing \(\qimp^\DBV\) in the expression of the drift, we obtain that, when \(Z^t \ge (\eps - \eps')n\) and \(\lambda^t \ge \lambda^{*, \DBV}(\eps n , s) + 7/8\), the drift of \(H\) is  
    \begin{align*}
        \EE[H^t - H^{t+1} \mid Z^t, \lambda^t] 
            &\le - \frac{1+s}{2s}d_2,
    \end{align*}
    i.e., a negative constant. We then conclude (up to a change of sign) as in the proof of~\ref{thm:itm:existence_threshold_1}: for \(\lambda^t\) to increase from \(\lambda^{*, \DBV}(\eps n, s) - 7/8\) and exceed \(\lambda^{*, \DBV}(\eps n , s) + 1\), while \(Z^t \ge (\eps - \eps') n\) it must be that \(-H^t\) has increased by \(\omega(\log n)\) despite negative drift. By Theorem~\ref{thm:NegativeDrift}, this event does not hold for a superpolynomial number of steps with high probability.

    % % We first show that as long as $Z^t \ge (\eps-\eps')n$, we have $\lambda^t \le \lambda^{*,\DBV}(\varepsilon n,  s) + 1$ for a superpolynomial number of steps. 
    
    % Without loss of generality, \(\eps'\) is small enough to apply  Lemma~\ref{lem:pimp-relations}~\ref{lem:itm:pimp-relations-2} and we have \(\lambda^{*, \DBV}((\varepsilon - \varepsilon')n, s) \le \lambda^{*, \DBV}(\eps n , s) + 1/4\).
    % As for case a)
    % % , by possibly decreasing $\eps'$, 
    % we can achieve that for all \(\lambda^t \ge \lambda^{*, \DBV}(\eps n , s) + 7/8 \ge \lambda^{*, \DBV}((\eps - \eps') n , s) + 5/8 \) and all \(Z^t = ( \eps \pm \eps')n\),
    % \begin{align*}
    %     \qimp^{\DBV}(Z^t,\lfloor\lambda^t\rceil) \ge \qimp^{\DBV}(( \eps - \eps') n, \lambda^{*,\DBV}( (\eps - \eps') n, s) + 1/8) > \tfrac{1}{s+1},
    % \end{align*}
    % % 
    % which implies a constant negative drift of $\lambda^t$. As in case a), with high probability $\lambda^t$ does not exceed $\lambda^{*, \DBV}(\varepsilon n, s)+1$ for a superpolynomial number of steps, as long as $Z^t \ge ( \eps -\eps')n$ is satisfied. 

    To conclude, we establish that it takes a superpolynomial number of steps for \(Z^t\) to drop\footnote{Mind the change of sign compared to the statement of Theorem~\ref{thm:NegativeDrift}.} from \(a = (\eps + \eps') n\) to \(b = (\eps - \eps')n\) while \(\lambda^t \le \lambda^{*, \DBV}(\eps n , s) + 1\) using the Negative Drift Theorem~\ref{thm:NegativeDrift}. Under the previous assumption on \(\lambda^t\), $Z^t$ has constant drift away from the optimum in the interval $(\eps \pm \eps')n$ by Lemma~\ref{lem:construction_inefficient_DBV}, i.e., condition~\ref{thm:itm:NegativeDrift-1} is satisfied. Since we may assume that \(|Z^t - Z^{t+1}| \le \log n\) for a superpolynomial number of steps by Lemma~\ref{lem:small-steps}, condition~\ref{thm:itm:NegativeDrift-2} is also satisfied for \(r = \log n\). As \(\ell = a - b = \Omega(n)\), the last condition~\ref{thm:itm:NegativeDrift-3} is also satisfied (recall that we assume \(n\) large enough). The Negative Drift Theorem~\ref{thm:NegativeDrift} then implies that with high probability $Z^t$ stays above $( \eps -\eps')n$ for a superpolynomial number of steps. This concludes the proof of \ref{thm:itm:existence_threshold_2}.
    \qed
\end{proof}

\section{Simulations}\label{sec:simulations}

This section aims at providing empirical support to our theoretical results. Namely, we show that there exist parameters $s$ and $F$ such that \onemax is optimized efficiently by the \saoclea, while \dynBV is not. Moreover, in the simulations we find that the claim also extends to non-dynamic functions, such as \BinVal and \binary\footnote{Defined as $\BinVal(x) = \sum_{i=1}^{n} 2^{i-1} x_i $ and $\binary(x) = \sum_{i=1}^{\lfloor n/2 \rfloor} x_i n + \sum_{i=\lfloor n/2 \rfloor + 1}^n x_i$.}. In all experiments, we set $n=1000$, the update strength $F=1.5$, and the mutation rate to be $1/n$. 
For this choice of update strength and mutation rate, Hevia Fajardo and Sudholt~\cite{hevia2021self} empirically found a threshold \(s_0 \approx 3.4\) for \OM, that is, the optimisation of \onemax is efficient when \(s < s_0\), and it is not when \(s > s_0\). For our experiments, we choose first \(s = 3\), which is close enough to the observed threshold, followed by \(s = 2\), further below the threshold.
Then we start the \saoclea with the zero string and an initial offspring size of $\lambda^{\text{init}} = 1$. The algorithm terminates when the optimum is found or after $500 n$ generations. The code for the simulations can be found at \url{https://github.com/zuxu/OneLambdaEA}.

\begin{figure}[ht]
    \centering
    \includegraphics[width=.7\textwidth]{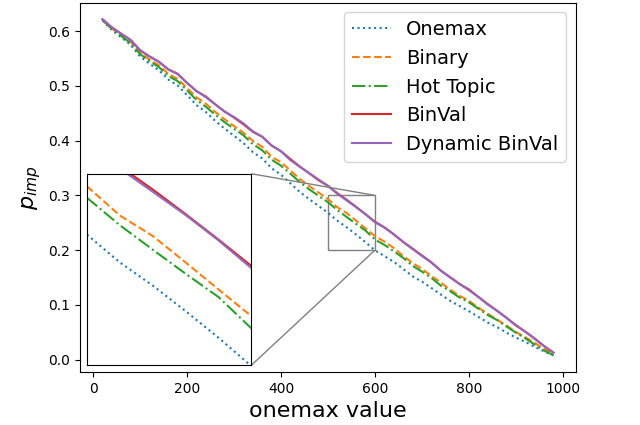}
    \caption{The probability of fitness improvement with a single offspring for search points with different \onemax values. Each data point in the figure is estimated by first sampling 1000 search points of the corresponding \onemax value, then sampling 100 offspring for each of the sampled search points, and calculating the frequency of an offspring fitter than its parent. The parameters of \hottopic~\citep{lengler2019general} are $L=100$, $\alpha=0.25$, $\beta=0.05$, and $\eps=0.05$.}
    \label{fig:improvement}
\end{figure}

\begin{figure}[H]
    \centering
    \includegraphics[width=1\textwidth]{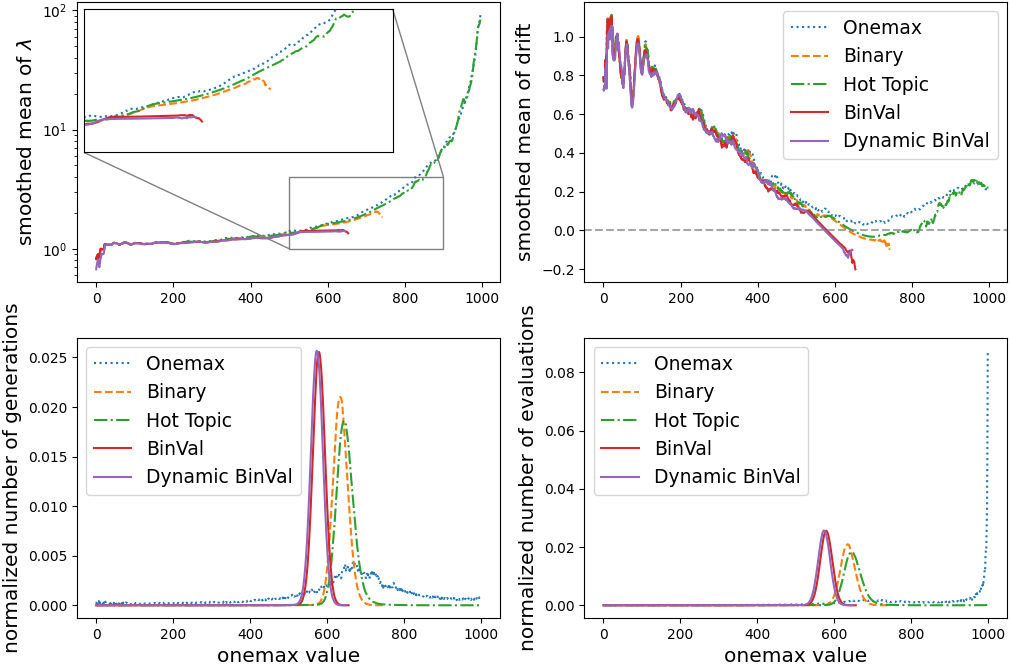}
    \caption{Smoothed average of $\lambda$, smoothed average drift, average number of generations, and average number of evaluations of the self-adjusting \oclea with $s=3$, $F=1.5$, and $c=1$ in 100 runs at each \onemax value when optimizing monotone functions with $n=1000$. The parameters of \hottopic~\citep{lengler2019general} are the same as the ones in Figure \ref{fig:improvement}. The average of $\lambda$ is shown in log scale. The average of $\lambda$ and the average drift are smoothed using a moving average over a window of size 15. The number of generations/evaluations is normalized such that its sum over all \onemax values is 1.}
    \label{fig:average3}
\end{figure}

\begin{figure}[!hb]
    \centering
    \includegraphics[width=1\textwidth]{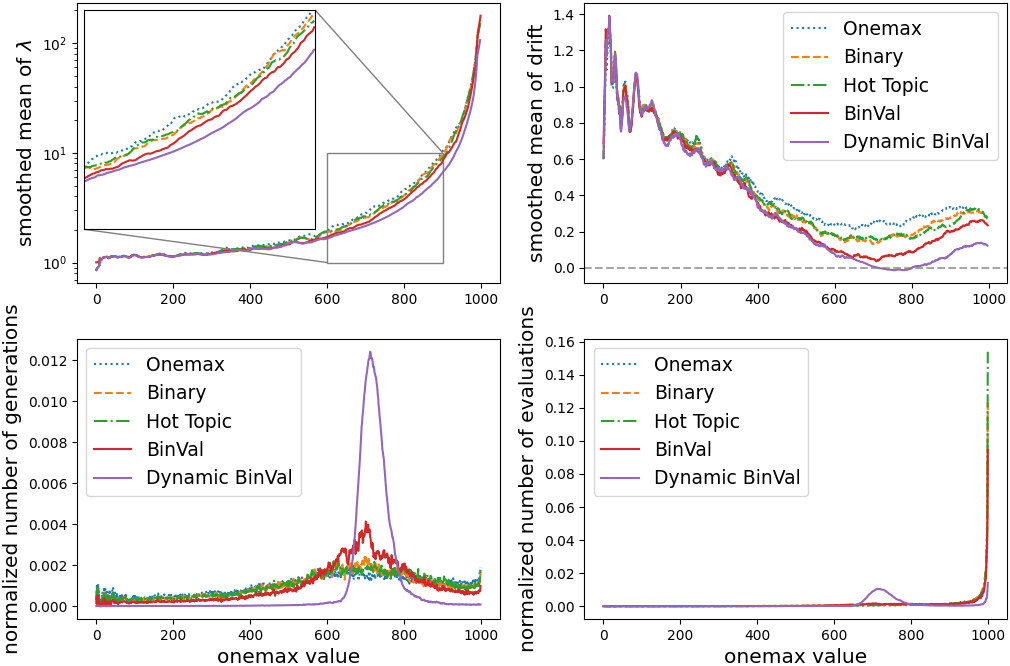}
    \caption{The same experiments as shown in Figure \ref{fig:average3} using $s=2$ instead of $s=3$.}
    \label{fig:average2}
\end{figure}

We first show that the improvement probability $p_{\text{imp}}$ of \onemax is the lowest among all considered monotone functions (Figure \ref{fig:improvement}), while it is highest for \dynBV and \binary (partly covered by the violet line of \dynBV, that is, the line closest to the top-right corner). Hence the fitness landscape looks hardest for \onemax with respect to fitness improvements. Therefore, to maintain a target success probability of $1/(1+s)$, more offspring are needed for \onemax, and the \saolea chooses a slightly higher $\lambda$ (first panel of Figures \ref{fig:average3} and \ref{fig:average2}, partly covered by the green line). This contributes positively to the drift towards the optimum, so that the drift for \onemax is higher than for the other functions (second panel). 
%do observe a negative correlation between $p_{\text{imp}}$ and $\lambda$, and a positive correlation between $\lambda$ and the drift.

% \begin{figure}[ht]
%     \centering
%     \includegraphics[width=.7\textwidth]{figures/improvement.png}
%     \caption{The probability of fitness improvement with a single offspring for search points with different \onemax values. Each data point in the figure is estimated by first sampling 1000 search points of the corresponding \onemax value, then sampling 100 offspring for each of the sampled search points, and calculating the frequency of an offspring fitter than its parent. The parameters of \hottopic~\citep{lengler2019general} are $L=100$, $\alpha=0.25$, $\beta=0.05$, and $\eps=0.05$.}
%     \label{fig:improvement}
% \end{figure}

\vfill

\begin{figure}[!hb]
    \centering
    \includegraphics[width=1\textwidth]{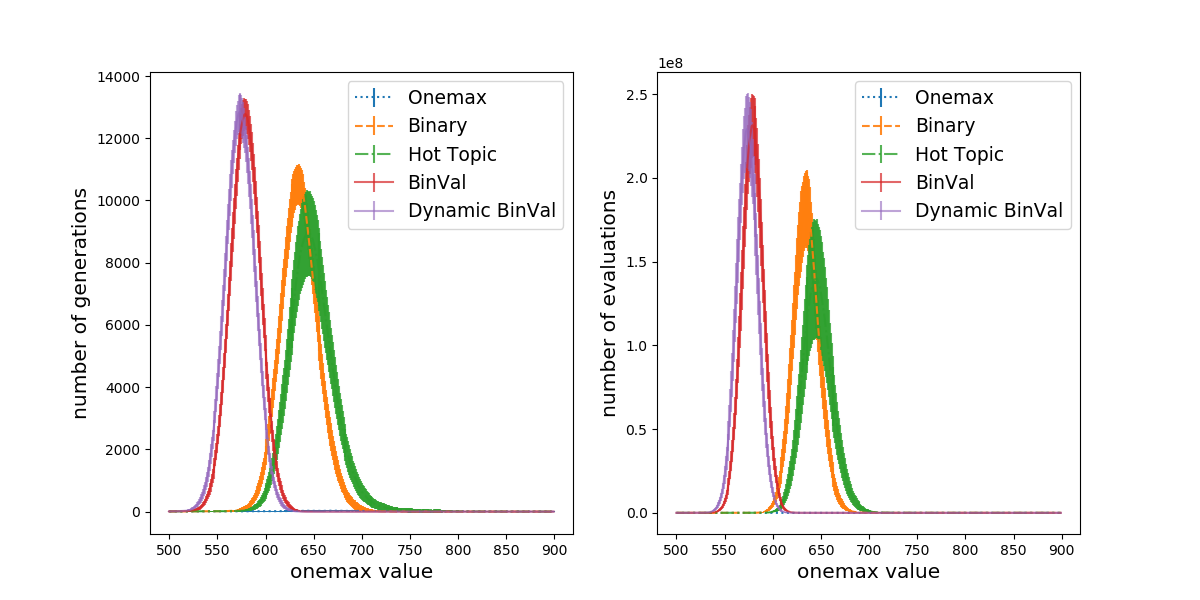}
    \caption{Average number of generations, and average number of evaluations of the self-adjusting \oclea with $s=3$, $F=1.5$, and $c=1$ in 100 runs plotted against \onemax values when optimizing monotone functions with $n=1000$. The parameters of \hottopic~\citep{lengler2019general} are the same as the ones in Figure \ref{fig:improvement}. The shaded areas indicate one standard deviation above and below the respective means.}
    \label{fig:deviation3}
\end{figure}

\begin{figure}[!htb]
    \centering
    \includegraphics[width=1\textwidth]{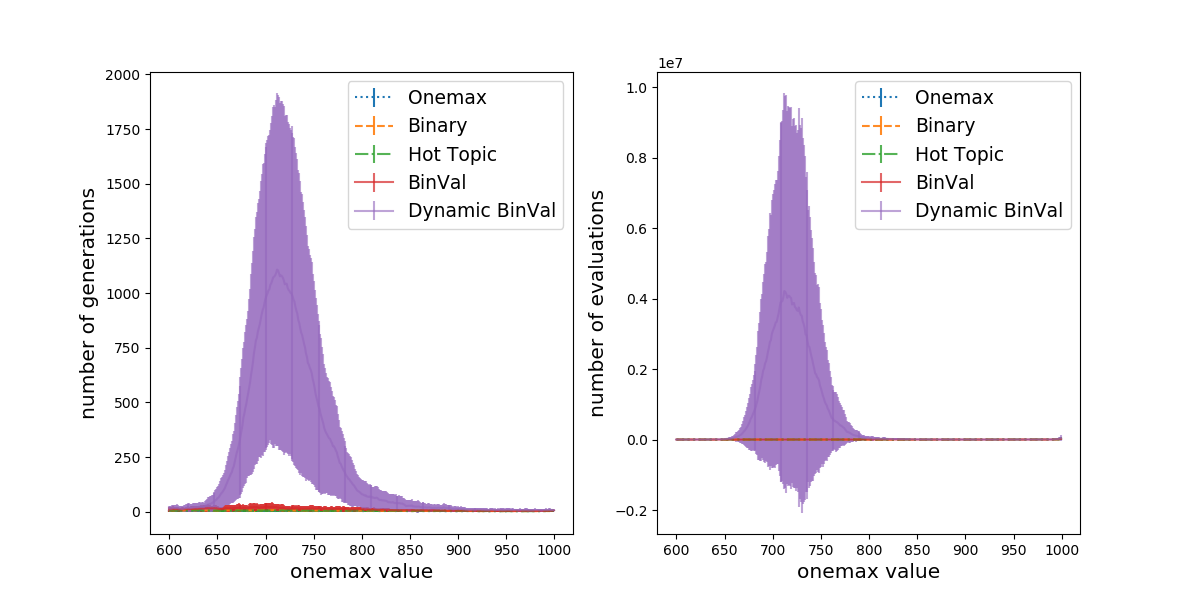}
    \caption{The same experiments as shown in Figure \ref{fig:deviation3} using $s=2$ instead of $s=3$.}
    \label{fig:deviation2}
\end{figure}

Figure~\ref{fig:average3} summarizes our main result of this section. Clearly, we observe that the \saoclea gets stuck on all considered monotone functions except \onemax when the number of one-bits in the search point is between $0.55n$ and $0.65n$. Although the algorithm spends a bit more generations on \onemax between $0.5n$ and $0.8n$ compared to the other parts, the optimum is found rather efficiently (not very pronounced in the figure though, since the number is normalized in order to show the proportion with which each \onemax value is attained). The reason is, all functions except \onemax have negative drifts somewhere within the interval $(0.58n, 0.8n)$. The drift of \onemax here is also small compared to the other regions, but remains positive. We also note that the optimum is also found for \hottopic~\citep{lengler2019general} despite its negative drift at $0.7 n$, probably because the drift is so weak that it can be overcome by random fluctuations. 

% \begin{figure}[ht]
%     \centering
%     % \includegraphics[width=1.11\textwidth]{figures/average_3.png}
%     \includegraphics[width=1\textwidth]{figures/average_3_cropped.png}
%     \caption{Smoothed average of $\lambda$, smoothed average drift, average number of generations, and average number of evaluations of the self-adjusting \oclea with $s=3$, $F=1.5$, and $c=1$ in 100 runs at each \onemax value when optimizing monotone functions with $n=1000$. The parameters of \hottopic~\citep{lengler2019general} are the same as the ones in Figure \ref{fig:improvement}. The average of $\lambda$ is shown in log scale. The average of $\lambda$ and the average drift are smoothed using a moving average over a window of size 15. The number of generations/evaluations is normalized such that its sum over all \onemax values is 1.}
%     \label{fig:average3}
% \end{figure}

As a comparison, we show in Figure \ref{fig:average2} the situation when $s$ is smaller. Due to a smaller value of $s$, all functions have positive drifts toward the optimum most of the time during the optimization progress, and the global optimum is found for all of them. As the 'hardest' function considered in our simulations, \dynBV has a lower drift compared to the other functions after $0.6 n$ and a slightly negative drift between $0.7 n $ and $0.8 n$, which leads to a peak in generations during this particular interval.

% \begin{figure}[ht]
%     \centering
%     % \includegraphics[width=1.11\textwidth]{figures/average_2.png}
%     \includegraphics[width=1\textwidth]{figures/average_2_cropped.png}
%     \caption{The same experiments as shown in Figure \ref{fig:average3} using $s=2$ instead of $s=3$.}
%     \label{fig:average2}
% \end{figure}

For added visual clarity, we have omitted standard deviation bounds from Figures~\ref{fig:average3} and 
\ref{fig:average2}. For a sense of how concentrated around their respective means the number of generations and evaluations are across several optimisation runs, we refer the reader to Figures~\ref{fig:deviation3} and \ref{fig:deviation2}. Note that those numbers are not normalized as in the previous figures, so the number of generations and evaluations from \onemax appears negligible compared to \dynBV. 

% \begin{figure}[ht]
%     \centering
%     \includegraphics[width=1\textwidth]{figures/deviation_3.png}
%     \caption{Average number of generations, and average number of evaluations of the self-adjusting \oclea with $s=3$, $F=1.5$, and $c=1$ in 100 runs plotted against \onemax values when optimizing monotone functions with $n=1000$. The parameters of \hottopic~\citep{lengler2019general} are the same as the ones in Figure \ref{fig:improvement}. The shaded areas indicate one standard deviation above and below the respective means.}
%     \label{fig:deviation3}
% \end{figure}

% \begin{figure}[ht]
%     \centering
%     % \includegraphics[width=1.11\textwidth]{figures/average_2.png}
%     \includegraphics[width=1\textwidth]{figures/deviation_2.png}
%     \caption{The same experiments as shown in Figure \ref{fig:deviation3} using $s=2$ instead of $s=3$.}
%     \label{fig:deviation2}
% \end{figure}

\section{Conclusion}\label{sec:conclusion}

The key insight of our work is that there are two types of ``easiness'' of a benchmark function, which need to be separated carefully\footnote{We note that there are other types of ``easiness'', e.g. with respect to a fixed budget.}. The first type relates to the question of how much progress an elitist hillclimber can make on the function. In this sense, it is well-known that \onemax is indeed the easiest benchmark among all functions with unique global optimum. However, a second type of easiness is how likely it is that a mutation gives an improving step. Here \onemax is not the easiest function.

Once those concepts are mentally separated, it is indeed not hard to see that \onemax is not the easiest function with respect to the second type. In this paper we have shown that \dynBV is easier (Lemma~\ref{lem:pimp-relations}), but we conjecture that this actually holds for many other functions as well, including static ones. This is backed up by experimental data, but we are lacking a more systematic understanding of which functions are hard or easy in this aspect.

We have also shown that the second type of easiness is \emph{relevant}. In particular, the \saolea relies on an empirical sample of the second type of easiness (aka the improvement probability) to choose the population size. Since the \saolea may make bad choices for too easy settings (of second type) if the parameter $s$ is set too high, it is important to understand how easy a fitness landscape can get. These easiest fitness landscapes will determine the range of $s$ that generally makes the \saolea an efficient optimizer. We have disproved the conjecture from~\cite{hevia2021arxiv,hevia2021self} that \onemax is the easiest function (of second type). 
As an alternative, we conjecture that the easiest function is the `adversarial' \dynBV, defined similarly to \dynBV with the exception that the permutation is not random, but chosen so that any \(0\)-bit is heavier than all \(1\)-bits. With this fitness function, any mutation in which at least one \(0\)-bit is flipped gives a fitter child, regardless of the number of \(1\)-bit flips, so it is intuitively convincing that it should be the easiest function with respect to fitness improvement.

%As an alternative, we conjecture that \dynBV is the easiest function, but we are not confident in our conjecture, and more research is needed to answer this question. \ML{For consistency with the conclusion of the other paper, I would say we conjecture that \emph{adversarial} \DBV is the easiest. (And explain ADBV in a few sentences.}

% \newpage 

\vfill

\subsubsection*{Acknowledgements.} We thank Dirk Sudholt for helpful discussions during the Dagstuhl seminar 22081 ``Theory of Randomized Optimization Heuristics'' and Mario Hevia Fajardo for sharing his simulation code for comparison. Marc Kaufmann and Xun Zou were supported by the Swiss National Science Foundation, grant number 200021\_192079 and grant number CR-SII5\_173721 respectively.

\bibliographystyle{apalike}
\bibliography{references}
\end{document}